\documentclass[journal,twoside,web]{ieeecolor}
\usepackage{generic}
\usepackage{cite}
\usepackage{amsmath,amssymb,amsfonts}
\usepackage{algorithmic}
\usepackage{graphicx}
\usepackage{algorithm}
\usepackage{hyperref}
\hypersetup{hidelinks=true}
\usepackage{textcomp}

\begin{document}
\title{Graph-Radiomic Learning (GrRAiL) Descriptor to Characterize Imaging Heterogeneity in Confounding Tumor Pathologies}
\author{Dheerendranath Battalapalli, Apoorva Safai, María Jaramillo, Hyemin Um,
Gustavo Adalfo Pineda Ortiz, Ulas Bagci, Manmeet Singh Ahluwalia, Marwa Ismail, and Pallavi
Tiwari \textbf{*}, \IEEEmembership{Member, IEEE}
\thanks{"This research was funded by NIH/NCI/ITCR: 1U01CA248226-01, NIH/NCI R01CA277728-01A1, NIH/NCI: 1R01CA264017-01A1, NIH/NCI: 3U01CA248226-03S1, DOD/PRCRP Career Development Award: W81XWH-18-1-0404, The Dana Foundation David Mahoney Neuroimaging Program, The V Foundation Translational Research Award, The Johnson \& Johnson WiSTEM2D Award, Musella Foundation Grant, R\&D Pilot Award, Departments of Radiology and Medical Physics, University of Wisconsin-Madison, and WARF Accelerator Oncology Diagnostics Award". }
\thanks{Dheerendranath Battalapalli, Apoorva Safai, María Jaramillo, Gustavo Adalfo Pineda Ortiz, Marwa Ismail, Hyemin Um, and Pallavi Tiwari are with the Department of Biomedical Engineering and Department of Radiology at University of Wisconsin-Madison, Madison, WI 53706 USA.(e-mail: battalapalli@wisc.edu; Safai@wisc.edu; jaramillogon@wisc.edu; gpinedaortiz@wisc.edu; ismail8@wisc.edu; hum3@wisc.edu; ptiwari9@wisc.edu).}
\thanks{\textbf{*} Pallavi Tiwari is with William S. Middleton Memorial Veterans Affairs (VA) Healthcare, Madison, WI, USA (e-mail: ptiwari9@wisc.edu)-\textbf{corresponding author}.}
\thanks{Ulas Bagci is with Department of Radiology, Northwestern University, Evanston, IL 60208 USA (e-mail: ulas.bagci@northwestern.edu.}
\thanks{Manmeet Singh Ahluwalia are with Miami Cancer Institute, Baptist Health South Florida, Miami, FL 33139 USA (e-mail: Mainak.Bardhan@baptisthealth.net; manmeetA@baptisthealth.net).}
}

\maketitle

\begin{abstract}

A significant challenge in solid tumors is the reliable distinction of confounding pathologies from malignant neoplasms on routine imaging scans. While radiomics approaches have attempted to capture surrogate markers of lesion heterogeneity on clinical imaging (CT, MRI), these approaches often aggregate feature values across the region of interest (ROI) and may not capture the complex spatial relationships across various intensity compositions in the ROI.  
We present a new Graph-Radiomic Learning (GrRAiL) descriptor for characterizing intralesional heterogeneity (ILH) on clinical MRI scans. GrRAiL involves (1) identifying clusters of sub-regions based on per-voxel radiomic measurements, followed by (2) computing graph-theoretic measurements that analyze the spatial associations of radiomic clusters. The weighted graphs constructed via the
GrRAiL descriptor capture the complexity of the higher-order
spatial relationships within the ROI, in an attempt to reliably characterize ILH and distinguish confounding pathologies from malignant neoplasms. To evaluate its efficacy and clinical feasibility, GrRAiL descriptor was evaluated on a total of $n=947$ subjects spanning 3 different clinical use-cases including distinguishing radiation-effects from tumor recurrence in glioblastoma ($n = 106$) and brain metastasis ($n = 233$), as well as distinguishing no+low- versus high-risk pancreatic intraductal
papillary mucus neoplasms (IPMNs) ($n = 608$). In a multi-institutional setting,  GrRAiL consistently outperformed state-of-the-art approaches (Graph Neural Networks (GNNs), textural radiomics, and intensity graph analysis). 
For the glioblastoma use-case, the cross validation (CV) and test accuracies obtained to classify tumor recurrence from pseudo-progression, were $89\%$ and $78\%$, respectively, with $>10\%$ improvement in test accuracy over comparative approaches. In the brain metastasis cohort, GrRAiL achieved CV and test accuracies of $84\%$ and $74\%$ in classifying tumor recurrence from radiation necrosis ($>13\%$ improvement over other approaches). Finally, when GrRAiL was used to classify between no+low- and high-risk IPMNs, CV and test accuracies of $84\%$ and $75\%$ were obtained, demonstrating $>10\%$ improvement over comparative approaches.

\end{abstract}


\section{Introduction}
\label{sec:introduction}
\IEEEPARstart{A} critical challenge in oncology is the frequent diagnostic ambiguity in differentiating between benign confounding pathologies and malignant neoplasms on clinical MRI scans \cite{verhaak2010integrated}. For instance, in neuro-oncology, the distinction between true tumor recurrence (TuR) and benign treatment-related effects (i.e. psuedoprogression (PsP), radiation necrosis (RN)) on post-treatment MRI scans, presents a well-known diagnostic dilemma \cite{zhou2022treatment} \cite{pope2018brain} \cite{derks2022brain}. TuR and benign treatment effects exhibit similar MRI visual characteristics, which often require invasive surgical interventions for a definitive diagnosis \cite{mitsuya2010perfusion} \cite{mayo2023radiation}. Similarly, in the screening for pancreatic cancer, the early identification of premalignant lesions represents a crucial strategy to improve patient prognosis. Despite an increased detection rate for incidental pancreatic cysts, their accurate classification into 'high-risk' and 'low-to-no-risk' categories remains a significant challenge \cite{chakraborty2017preliminary}.   

\begin{figure*}[t!]
\centering
\includegraphics[width=1\textwidth]{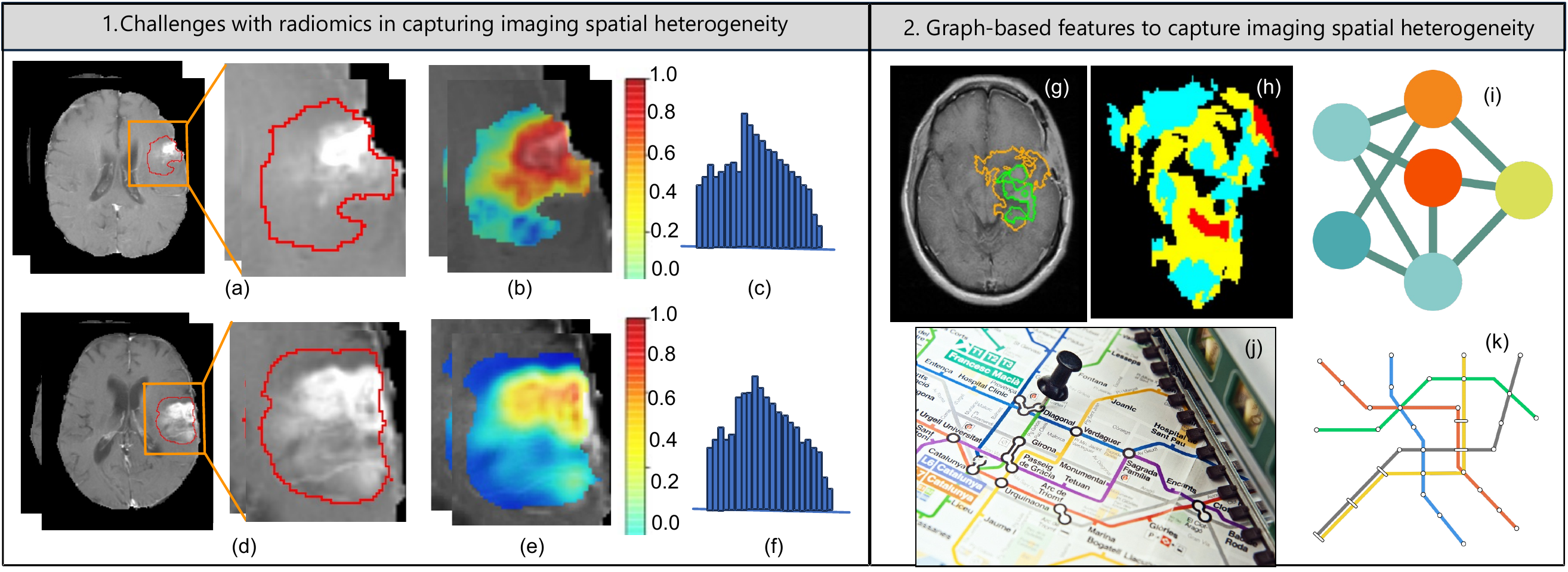}
\caption{\label{fig:res1}(a, d) demonstrate Gd-T1w MRI scans for TuR and PsP  in glioblastoma studies, respectively. (b, e) highlight the corresponding GLCM entropy expression maps, while (c, f) are the histogram plots from subjects with confirmed TuR and PsP, respectively. It is worth noting that while the spatial distribution of the radiomic feature expressions in (b) and (e) is distinct, the histogram distribution of the features in (c) and (f) appears similar, which could potentially contribute to misclassification of confounding pathologies (i.e., TuR versus PsP). (h) represents a clustered GLCM radiomics map (with 3 clusters) corresponding to a glioblastoma Gd-T1w MRI scan, and the analogy of capturing spatial interactions across radiomic clusters via a graphical representation (i) with that of a subway map shown in (j) and (k). }
\end{figure*}
Fundamentally, malignant lesions are characterized by a complex and distinct microenvironment organization with different tissue types disbursed through the lesion, in contrast to benign lesions \cite{gerlinger2012intratumor} \cite{sottoriva2013intratumor}. Thus, the ability to analyze the subtle organizational structure of tumors and their underlying heterogeneity could enable reliable discrimination of benign confounding pathologies from malignant lesions. This noninvasive characterization could also significantly impact clinical decision making, thus mitigating the need for unnecessary biopsies and treatments in patients with benign conditions.

Several approaches have attempted to capture surrogate markers of lesion heterogeneity on imaging (e.g. MRI, CT) to distinguish between confounding pathologies, including (1) radiomics that encompasses intensity-based, textural, and topological characteristics of the lesion~\cite{zhou2014radiologically,zhou2017identifying,wu2016robust, sala2017unravelling, zhou2018radiomics, corral2019deep, lalonde2019inn, hussein2019lung}, 
and (2) deep learning models that extract complex features directly from the images without the need for semantic segmentation, including convolutional
neural networks (CNN) and graph neural networks (GNN). While deep learning approaches may demonstrate advantages over radiomics with larger training cohorts, DL approaches
tend to be `opaque' with regard to feature interpretability.   

Among radiomic techniques, the Gray Level co-occurrence Matrix (GLCM) quantifies textural variations within the region of interest (ROI) at a voxel-level, correlating imaging heterogeneity with tumor aggressiveness across cancers \cite{van2020radiomics, horvat2018mr, mccague2023introduction,ismail2022radiomic, ismail2018shape, battalapalli2023fractal, yang2015discrete,lambin2012radiomics, a2016polypharmacology, aerts2014decoding, ibrahim2021radiomics}.  However, a significant methodological limitation with radiomic features, including GLCM, lies in the subsequent consolidation of the texture maps into aggregated feature metrics (e.g., mean, mode, kurtosis) \cite{antunes2022radiomic}. Although these aggregated values seek to capture the `image-based heterogeneity' associated with disease aggressiveness, they inherently lose the spatial information regarding the organization of different tissue types in a lesion and may not represent the distinct pathological variabilities  within the same lesion \cite{antunes2022radiomic}. 

For example, consider the radiomic expression maps in Figure 1, which represent textural patterns derived from GLCM feature maps for two confounding pathologies (TuR, TrE). Visual inspection of these expression maps reveals discernible spatial differences in lesion organization, reflected by variations in the color distribution on the GLCM entropy expression heatmaps from the segmented ROI. In a way, these radiomic feature expression maps (Figure 1(h)) are analogous to a subway map (Figure 1(j), (k)). The radiomic maps often represent overlapping regions where different cell populations interact, influencing heterogeneity, while subway maps demonstrate spatial intersections across transfer stations with multiple lines converging.  Unfortunately, both machine learning (ML) and DL classifiers may not explicitly encode the spatial interactions of the 'sub-populations' (i.e. clusters) within the ROI for reliable distinction of confounding pathologies from malignant lesions. Consequently, there exists an opportunity to advance beyond aggregated feature values, by investigating intensity composition and complex interrelationships within the tumor microenvironment, at the spatial level. Such an approach could enable a comprehensive characterization of the lesion's inherent heterogeneity and improve the distinction of confounding pathologies from malignant tumors/cysts.

Recently, some studies have attempted to go beyond conventional radiomic analysis to explore structural connectivity within the lesion and how it impacts heterogeneity patterns and tumor behavior. For instance, \cite{lee2023association,cao2023multidimensional} explored the utility of graph theory approaches in characterizing primary and metastatic brain tumor heterogeneity by quantifying spatial interactions within intratumoral sub-regions. In addition, \cite{kocevar2016graph} presented an automated pipeline that leverages structural connectivity graphs (derived from T1 and DTI data) and global graph metrics to accurately classify multiple sclerosis (MS) patients into four clinical subtypes. Similarly, our group developed a method called RADIomic Spatial TexturAl descripTor (RADISTAT) to analyze the spatial interactions within the tumor regions\cite{antunes2022radiomic}, via categorizing voxels into sub-regions of low, intermediate, and high intensity using pairwise interactions, to analyze the tumor's complex structure, evaluating its efficacy in the context of brain and rectal cancer post-treatment response assessment on imaging. 

In this work, we present a new Graph Radiomic Learning approach (GrRAiL) to comprehensively characterize imaging heterogeneity by (1) identifying clusters of sub-regions based on per-voxel radiomic measurements, followed by (2) computing graph-theoretic measurements that analyze the spatial associations of the radiomic clusters. The weighted graphs constructed via the GrRAiL descriptor capture the complexity of the higher-order spatial relationships within the lesion microenvironment to differentiate between confounding pathologies.  A preliminary implementation of 3D GrRAiL was previously presented in \cite{Battalapalli2024}. Our current work builds on the initial work to include a detailed analysis of the GrRAiL algorithm with extended parameter sensitivity analysis. Additionally, we perform extensive comparisons of GrRAiL with state-of-the-art approaches in the context of clinically relevant problems in brain and pancreatic lesions. We summarize our key contributions as follows:
\begin{itemize}
    \item Capturing the spatial organization of radiomic patterns within the lesion environment on structural MRI scans through clustering radiomic expression maps and generating graph theoretic features that characterize the spatial interactions of the radiomic patterns within the lesion. GrRAiL descriptor thus quantifies the spatial connectivity of 'sub-populations' within the lesion (analogous to a subway system),  and their influence on disease malignancy.
    
    \item Evaluating GrRAiL's efficacy  across clinical challenges in: (a) differentiating tumor recurrence (TuR) from radiation necrosis (RN) in metastatic brain tumors after radiotherapy on a multi-institutional cohort ($n=233$ studies), 
(b) distinguishing TuR from pseudo-progression (PsP) in glioblastoma on a multi-institutional cohort ($n=106$ studies), and
(c) Classifying Intraductal Papillary Mucinous Neoplasms (IPMNs), the most common pancreatic cysts, into no-risk + low-risk versus high-risk on a multi-institutional study comprising T2w MRI scans from $n=608$ studies. 
\end{itemize}

\section{METHODOLOGY}

\begin{figure*}[t!]
\centering
\includegraphics[height=20cm, width=1.0\textwidth]{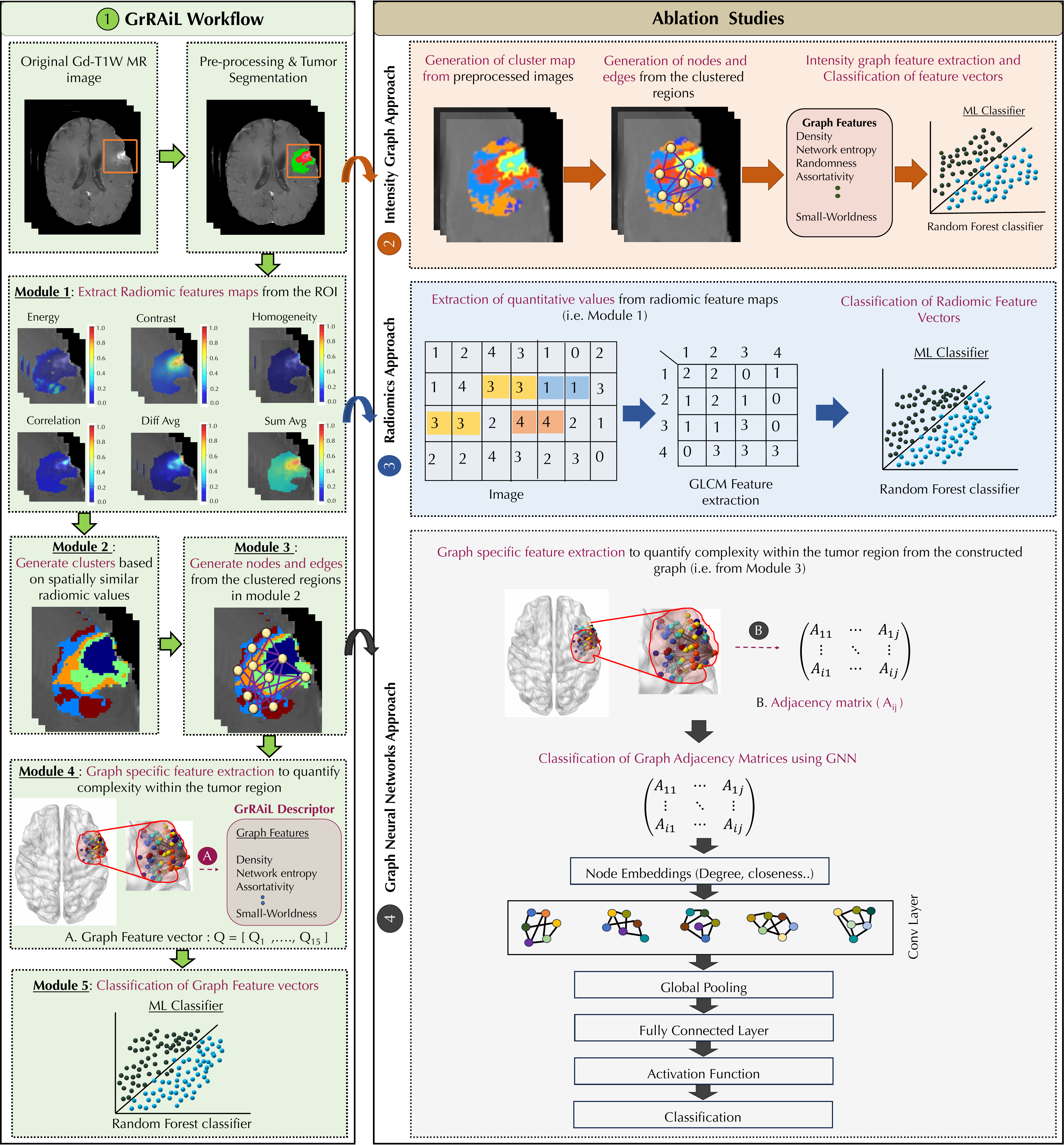}
\caption{\label{fig:res1} \textbf{Workflow illustrating the GrRAiL framework and accompanying ablation studies. GrRAiL Framework}: (1) Radiomic feature extraction: Compute voxel‐wise gray‐level co-occurrence matrix (GLCM) feature maps within the segmented region of interest. Heatmap overlay (red = high expression, blue = low). (2) Subregion clustering: Apply Gaussian Mixture Modeling (GMM) to each radiomic map; select optimal cluster count via Bayesian Information Criterion (BIC). (3) Graph construction: Represent each cluster centroid as a node; define edge weights using Earth Mover’s Distance (EMD) between node pairs. (4) Graph‐theory feature extraction: Derive weighted graph metrics (e.g., node degree, clustering coefficient, path length) to quantitatively characterize spatial heterogeneity. (5) Classification: Perform feature selection on the graph‐theory metrics and train a classifier to distinguish confounding pathologies. \textbf{Ablation Studies}:   \textit{Intensity graph approach}: Apply GMM clustering directly on the raw ROI intensities (bypassing GLCM maps), construct graphs from those clusters, extract graph‐theory features as in Module 4, and classify via Random Forest. \textit{Radiomics approach}: Quantitative values were extracted from Module 1 GLCM feature maps were fed directly into a Random Forest classifier for classification.  
\textit{Graph neural network approach}: Follow the same clustering and graph‐construction pipeline (as in GrRAiL pipeline), but instead of extracting hand‐crafted graph metrics, input the full graph adjacency and node‐attribute matrices into a Graph Neural Network for direct classification.}
\end{figure*}

\subsection{Image Notation}
We define a GLCM feature expression map as \( I = (K, f) \), where \( K \) denotes a spatial grid composed of pixels \( k \in \mathbb{R}^3 \). Each pixel \( k \in K \) is associated with a specific value of the radiomic feature \( f(k) \). Figure~2 illustrates an overview of the GrRAiL framework. The notations and acronyms used throughout this paper are summarized in Table~I.

\subsection{GLCM Feature map Extraction}

We extracted GLCM feature maps from the entire tumor region.  A total of 13 GLCM feature maps $I_1, \ldots, I_{13}$ were derived: energy, entropy, contrast, correlation, homogeneity, sum average, sum variance, sum entropy, difference entropy, difference average, difference variance, information measure of correlation 1 (ICM1), and information measure of correlation 2 (ICM2) \cite{haralick1973textural}.

\begin{table*}[!t]
  \caption{List of the Notations and Acronyms Used in This Paper}
  \label{tab:notations}
  \centering
  \footnotesize
  \renewcommand{\arraystretch}{1.1}
  \begin{tabular}{|l|l|l|l|}
    \hline
    \textbf{Notation} & \textbf{Description} 
      & \textbf{Notation} & \textbf{Description} \\ 
    \hline
    $I$                    & Radiomic feature expression map $(K,f)$  
      & $\hat{I}$             & Clustered radiomic feature expression map $(K,g)$      \\
    \hline
    $K$                    & Spatial grid in $\mathbb{R}^3$  
      & $k$                   & Voxel in $K$                                           \\
    \hline
    $f(k)$                 & Radiomic feature value at voxel $k$  
      & $g(k)$                & Average radiomic feature value within the cluster of $k$ \\
    \hline
    $\hat{K}_{{u}}$        & Set of voxels in the ${{u}}$-th cluster  
      & ${{u}}$                   & Number of clusters (via BIC)                           \\
    \hline
    $M$                    & Total number of GMM components  
      & $\pi_{{u}}$               & Prior probability of component ${{u}}$                     \\
    \hline
    $\mu_{{u}}$                & Mean vector of Gaussian component ${{u}}$  
      & $\Sigma_{{u}}$            & Covariance matrix of component ${{u}}$                     \\
    \hline
    $\mathcal{N}(x_k | \mu_{{u}}, \Sigma_{{u}})$
                           & Gaussian PDF for $x_k$ under component ${{u}}$  
      & $P({{u}}\mid x_k)$        & Posterior probability that voxel $k$ belongs to ${{u}}$    \\
    \hline
    $C_{{u}}$                  & Centroid coordinates of cluster ${{u}}$  
      & $w_{ij}$              & Edge weight (EMD) between nodes $i$ and $j$            \\
    \hline
    $G=(V,E)$              & Undirected graph of centroids and edges  
      & $A_{ij}$              & Adjacency matrix entry (1 if edge $(i,j)$ exists)     \\
    \hline
    $Q=[Q_1,\dots,Q_{15}]$ & Global graph-theoretic feature vector  
      & $\mathbf{x}_k$        & Intensity-feature vector at voxel $k$                  \\
    \hline
    $I_i,\;i=1,\dots,13$   & The $i$-th GLCM feature map (energy, entropy, …, ICM2)  
      & ROI                   & Region of Interest                                     \\
    \hline
    \multicolumn{4}{|c|}{\textbf{Acronyms}} \\
    \hline
    \textbf{Acronym} & \textbf{Full Form} 
      & \textbf{Acronym} & \textbf{Full Form} \\ 
    \hline
    GLCM   & Gray-Level Co-occurrence Matrix  
      & GNN     & Graph Neural Network                                   \\
    \hline
    GMM    & Gaussian Mixture Model  
      & GIN     & Graph Isomorphism Network                              \\
    \hline
    BIC    & Bayesian Information Criterion  
      & GAT     & Graph Attention Network                                \\
    \hline
    RAG    & Region Adjacency Graph  
      & GCN-JK  & GCN with Jumping Knowledge                             \\
    \hline
    EMD    & Earth Mover’s Distance  
      & SOTA    & State-Of-The-Art                                       \\
    \hline
    HIPAA  & Health Ins.\ Portability \& Accountability Act  
      & T1w     & T1-weighted (MRI)                                      \\
    \hline
    IRB    & Institutional Review Board  
      & PsP     & Pseudo-progression                                     \\
    \hline
    TuR    & Tumor Recurrence  
      & RN      & Radiation Necrosis                                      \\
    \hline
    IPMN   & Intraductal Papillary Mucinous Neoplasm  
      & T2w     & T2-weighted (MRI sequence)                             \\
    \hline
    MRI    & Magnetic Resonance Imaging  
      & PanSegNet & Pancreas Segmentation Network                         \\
    \hline
    ITK-SNAP & Insight Segmentation and Registration Toolkit – SNAP  
      & CystX   & Cystic Neoplasm eXchange dataset                       \\
    \hline
    NYU    & New York University Langone Health  
      & MCF     & Mayo Clinic Florida                                    \\
    \hline
    NWU    & Northwestern University  
      & AHN     & Allegheny Health Network                               \\
    \hline
    MCA    & Mayo Clinic Arizona  
      & IU      & Istanbul University Faculty of Medicine                \\
    \hline
    EMC    & Erasmus Medical Center  
      &         &                                                        \\
    \hline
  \end{tabular}
\end{table*}

\subsection{{Clustering of Radiomic Feature Expression Maps}}

The clustering of $I$ is performed using a modified Gaussian Mixture Model (GMM) algorithm, to derive $u$ clusters \cite{zhao2021anomaly}.
The objective is to robustly delineate spatially contiguous regions in the radiomic expression map based on similar-valued voxels in $\mathbb{R}^3$, in an unsupervised manner. Our modified GMM approach assumes that the voxel intensities are generated from a finite mixture of Gaussian distributions, each symbolizing a unique cluster. 
The probability that a voxel $k$ with intensity vector $x_k$ belongs to cluster $u$ is defined by the following equation:

\begin{equation}
P(u \mid x_k) = \frac{\pi_{u} \mathcal{N}(x_k \mid \mu_{u}, \Sigma_{u})}{\sum_{i=1}^M \pi_i \mathcal{N}(x_k \mid \mu_i, \Sigma_i)},
\end{equation}

where \(\pi_{u}\) denotes the prior probability of cluster \(u\), and \(\mathcal{N}(x_k | \mu_{u}, \Sigma_{u})\) is the Gaussian probability density function with mean \(\mu_{u}\) and covariance \(\Sigma_{u}\) for the \({u}^{th}\) component. To determine the most suitable number of clusters for each individual radiomic expression map, we employ the Bayesian Information Criterion (BIC). 
It is worth noting that the number of clusters (\({u}\)) resulting from this step can vary as a function of the unique expression values in each \textit{I}. Based on the GMM clustering, \textit{I} is quantized to obtain a cluster map $\hat{I} = (K,g)$, where every \(k \in \hat{K}_{{u}} \subset K\), \(g(k)\) is the average radiomic feature value within cluster $\hat{K}_{{u}}$.

\subsection{Graph Construction}

Our framework is based on quantifying the clustered regions of radiomic expression levels in $\hat{I}$, in order to capture the spatial interactions among radiomic features within the constructed clusters. Considering the complex interactions within $\hat{I}$, we constructed a graph from it. Specifically, an undirected adjacency graph $G = (V, E)$ was defined, where $V$ is the set of nodes representing the centroid $C$ in each cluster \({u}\) within $\hat{I}$, and $E$ is the set of edges between the nodes in $V$. The edges in $G$ were defined using the Earth mover's distance (EMD) \cite{zhang2022deepemd} to calculate the weight between any two nodes $v_i$ and $v_j$, which can be represented as:
\begin{equation}
w_{ij} = \text{EMD}(C_{v_i}, C_{v_j})
\end{equation}
We denote the adjacency matrix of $G$ as $A = [A_{ij}]$ where $A_{ij}=1$ if there exists an edge $(v_i, v_j) \in E$ and $A_{ij} = 0$ otherwise.

\subsection{{Construction of GrRAiL descriptor}}

From the resulting $G$, the GrRAiL descriptor is obtained for every subject by computing the global features $\textbf{Q} = [Q_1, \dots, Q_{N}]$ for each radiomic map $I_1, \ldots, I_{13}$, where $N = 15$ is the total number of graph-theory based features extracted from the constructed graphs to quantify the structural properties of G. 
All global graph features are then concatenated across the extracted radiomic expression maps per subject.  The corresponding features and their description are provided in Table II.

\begin{algorithm}
\caption{Computation of GrRAiL Descriptor}
\label{alg:GrRAiL}
\begin{algorithmic}[1]
\renewcommand{\algorithmicrequire}{\textbf{Input:}}
\renewcommand{\algorithmicensure}{\textbf{Output:}}

\REQUIRE Radiomic feature expression map $I = (K, f)$, where $K$ is the spatial grid and $f$ is the feature.
\ENSURE  Global graph‐theory feature vector $\mathbf{Q}$

\textit{Initialization} :
\STATE Choose radiomic feature map $I$ from $I_1,\ldots,I_{13}$ for graph extraction.

\textit{Clustering Process} :
\STATE Generate ${{u}}$ clusters in $K$ using GMM, where $\hat{K}_{{u}} \subset K$, ${{u}} \in \{1,2,3,4,5\}$, and apply BIC to select optimal $u$.

\textit{Centroid Computation} :
\FOR{each cluster ${{u}}$ in $\hat{K}_{{u}}$}
    \STATE Compute centroid voxel components:

    \quad $C^x_{{u}} = \dfrac{1}{|\hat{K}_{{u}}|}\sum_{v_i \in \hat{K}_{{u}}} x_i$
     \quad $C^y_{{u}} = \dfrac{1}{|\hat{K}_{{u}}|}\sum_{v_i \in \hat{K}_{{u}}} y_i$
    \quad $C^z_{{u}} = \dfrac{1}{|\hat{K}_{{u}}|}\sum_{v_i \in \hat{K}_{{u}}} z_i$
\ENDFOR

\textit{Graph Construction} :
\STATE Generate edges using the EMD method to connect adjacent centroid voxels in $\hat{K}_{{u}}$.
\STATE Construct the weighted Region Adjacency Graph (RAG) $G = (V, E)$ from $\hat{K}_{{u}}$ and save adjacency matrix $A_{i,j}$.

\textit{Graph Metrics Extraction} :
\STATE Compute graph metrics $Q_1, \dots, Q_{15}$ (such as modularity, average path length, diameter, etc.).
\STATE Obtain graph features as a $1 \times N$ vector $\mathbf{Q}$ as described.

\textit{Save Output} :
\STATE Save the metrics to a file.

\RETURN $\mathbf{Q}$

\end{algorithmic}
\end{algorithm}

\begin{table}[h!]
\centering
\caption{Graph-Based Features and Their Descriptions}
\label{tab:graph_features}
\begin{tabular}{|p{3cm}|p{5cm}|}
\hline
\textbf{Feature} & \textbf{Description} \\ \hline
Size & Number of nodes in the graph. \\ \hline
Density & Proportion of possible edges that are actual edges. \\ \hline
Diameter & Maximal distance between any pair of nodes. \\ \hline
Average Shortest Path Length & Average distance across all pairs of nodes. \\ \hline
Clustering Coefficient & 
\parbox{5cm}{\[
C = \frac{3 \times \text{number of triangles in the graph}}{\text{number of connected triplets of nodes}}
\] 
Measures the degree of clustering among the nodes.} \\ \hline
Modularity & Strength of division of a network into modules or clusters. \\ \hline
Small-worldness & 
Compares the clustering coefficient to the characteristic path length relative to a random graph. \\ \hline
Connected Components & Number of connected subgraphs in the graph. \\ \hline
Assortativity & Correlation between node degrees at either end of an edge. \\ \hline
Radius & The minimum eccentricity of any node, where eccentricity is the greatest distance between a node and any other node. \\ \hline
Global Efficiency & The average inverse shortest path length, measuring how efficiently information is exchanged over the network. \\ \hline
Network Entropy & Measures the randomness or disorder within the network structure. \\ \hline
Number of Hubs & Number of nodes with significantly higher degrees compared to other nodes, acting as central points in the network. \\ \hline
Randomness Value & Quantifies the unpredictability of the network's structure. \\ \hline
Network Resilience & The ability of the network to maintain connectivity after the removal of nodes or edges. \\ \hline
\end{tabular}
\end{table}

\section{Experimental Design}

\begin{table*}[htbp]
  \centering
  \caption{Summary of Patient Characteristics, Imaging Parameters, and Outcome Groups in Discovery and Validation Cohorts}
  \label{tab:data}
  \resizebox{\textwidth}{!}{%
    \begin{tabular}{|l|c|c|c|c|c|c|}
      \hline
      & \multicolumn{2}{c|}{\textbf{Metastatic Brain Tumor}} & \multicolumn{2}{c|}{\textbf{Glioblastoma}} & \multicolumn{2}{c|}{\textbf{IPMN}} \\
      \hline
      & \textbf{Train (n=157)} & \textbf{Test (n=76)} & \textbf{Train (n=60)} & \textbf{Test (n=46)} & \textbf{Train (n=509)} & \textbf{Test (n=99)} \\
      \hline
      \multicolumn{7}{|l|}{\textbf{Patient Characteristics}} \\
      \hline
      Sex, men:women:n/a & 29:38:90 & 42:34 & 39:21 & 30:16 & 232:277 & 48:51 \\
      Age, mean (range), years & 56.85 (44--67) & 64.5 (55--74) & 60.6 (26--74) & 55.6 (25--76) & 65 (52--78) & 60 (46--74) \\
      \hline
      \multicolumn{7}{|l|}{\textbf{Imaging Parameters}} \\
      \hline
      MRI Protocol & Gd-T1w & Gd-T1w & Gd-T1w & Gd-T1w & T2w & T2w \\
      In-plane resolution (mm) & 1.00 & 1.00 & 1.00 & 1.00 & 1.00 & 1.00 \\
      Slice thickness (mm) & 0.8--10 & 1--2.40 & 1.00 & 1.00 & 4.5--4.6 & 5.6--6.1 \\
      Repetition time (msec) & 7.6--2200 & 7.7--2100 & 263--764 & 263--850 & NA & NA \\
      Echo time (msec) & 2.45--23 & 2.84--6 & 5--20 & 5--20 & --- & --- \\
      3T & 54 & 33 & 0 & 0 & NA & NA \\
      1.5T & 95 & 43 & 60 & 46 & NA & NA \\
      1T & 8 & 0 & 0 & 0 & NA & NA \\
      \hline
      \multicolumn{7}{|l|}{\textbf{Lesion Pathology}} \\
      \hline
      Tumor Recurrence & 112 & 38 & 37 & 33 & -- & -- \\
      Treatment Effects & 134 & 48 & 23 & 13 & -- & -- \\
      No risk + Low risk & -- & -- & -- & -- & 394 & 72 \\
      High risk         & -- & -- & -- & -- & 115 & 27 \\
      \hline
    \end{tabular}%
  }
  \vspace{1ex}
  \begin{flushleft}
    \textit{* In metastatic brain tumour studies, a single patient can have more than one lesion.
    Lesion-pathology counts are per scan. See data description for details.}
  \end{flushleft}
\end{table*}

\subsection{{Data Description}}
Our retrospective data curation across different use-cases was granted a waiver of informed consent by the institutional review board and was performed in compliance with Health Insurance Portability and Accountability Act (HIPAA) regulations. This multi-institutional retrospective study was approved by the University of Wisconsin–Madison Institutional Review Board: Glioblastoma and Metastasis cohorts (Approval No. 2023-0057) and IPMN cohort (Approval No. 2025-1191). The requirement for informed consent was waived for this retrospective analysis of de-identified data. A summary of patient characteristics, imaging parameters, and associated patient outcome groups across the 3 cohorts, is summarized in Table III. 

1) \textit{Glioblastoma Cohort:}
Our retrospective study consisted of MRI scans of $106$ patients: $46$ studies from Dana-Farber Cancer Center and $60$ studies from the Cleveland Clinic. University of Wisconsin–Madison Institutional Review Board (2023-0057); informed consent was waived for this retrospective analysis. The studies from Cleveland Clinic (TuR: $n=38$, PsP: $n=22$) were used as the training set, while those from the Dana-Farber Cancer Center (TuR: $n=33$, PsP: $n=13$) served as the test set. Glioblastoma patients who underwent chemoradiation treatment following the Stupp protocol and presented with a suspicious enhancing lesion within three months of treatment completion were included in the study. Each patient had a post-treatment Gadolinium-enhanced T1w MRI (Gd-T1w MRI), along with histological reports available for disease confirmation \cite{stupp2005radiotherapy}. To address known intensity discrepancies, all brain T1w MRI scans were processed for bias field correction, skull-stripping, and intensity normalization. The manual segmentations for enhancing lesions were obtained on every MRI slice that displayed more than 5 mm of rim enhancement, via consensus across two collaborating radiologists to minimize inter-reader variability. Ground truth definition for PsP or TuR was based on histologic confirmation via surgical resection or multiple biospies.

2) \textit{Metastatic Brain Tumor Cohort:}
We collected $n=233$ patient MRI studies of metastatic brain tumor patients: $n=109$ from Cleveland Clinic, $n=48$ from University Hospitals (Cleveland), and $n=76$ from the University of Wisconsin-Madison. Of the $n=233$ studies, some metastasis patients had more than one tumor, and thus we evaluated a total of $332$ lesions across $n=233$ studies.  For the training set, $157$ studies from Cleveland Clinic and the University Hospitals at Cleveland were used, comprising $246$ lesions: Cleveland Clinic (TuR: $n=74$, RN: $n=86$) and University Hospitals at Cleveland (TuR: $n=38$, RN: $n=48$). For the test set, $76$ studies from the University of Wisconsin–Madison Hospital were used, comprising $86$ lesions (TuR: $n=38$, RN: $n=48$). The inclusion criteria were: (1) patients treated with surgery and radiation therapy, (2) suspicious lesions on follow-up MRI before second surgical resection, (3) availability of surgically resected histopathological sections used for confirmation of disease or treatment effects, and (4) availability of post-contrast T1w, T2w, and FLAIR MRI scans. Patients were excluded if (1) there were any artifacts in the MRI scans, and (2) lesions were less than 5 mm in the largest axis. Patients receiving planned concurrent stereotactic radiosurgery (SRS) or whole-brain radiotherapy (WBRT) with immune checkpoint inhibitors (ICI) were included \cite{gondi2022radiation}. 

3) \textit{IPMN Cohort:}
Our retrospective multicenter IPMN study included $n=608$ patients with MRI scans (with both T1w and T2w) from seven centers. 
Participating centers in this study were the following: New York University (NYU) Langone Health ($\emph{n} = 147$), Mayo Clinic Florida (MCF) ($\emph{n} = 125$), Northwestern University (NWU) ($\emph{n} = 175$), Allegheny Health Network (AHN) ($\emph{n} = 12$), Mayo Clinic Arizona (MCA) ($\emph{n} = 15$), Istanbul University Faculty of Medicine (IU) ($\emph{n} = 62$), and Erasmus Medical Center (EMC) ($\emph{n} = 72$). Following pathological confirmation, cases were split across three categories: no-risk ($\emph{n} = 147$ cases), low-risk IPMN ($\emph{n} = 319$ cases), and high-risk IPMN ($\emph{n} = 142$ cases). Clinical information and MRI specifications of the scans are summarized in Table III.
Of the $608$ total patients, $509$ from IU ($\emph{n} = 62$), MCF ($\emph{n} = 125$), NWU ($\emph{n} = 175$), and NYU ($\emph{n} = 147$) were used for training, whereas the remaining $99$ from AHN ($\emph{n} = 12$), MCA ($\emph{n} = 15$), and EMC ($\emph{n} = 72$) served as the test set. 
This training–testing procedure was performed for no-risk+ low-risk vs high-risk classification task.
To segment the pancreas, we developed a semiautomatic method in two stages. In the first round, we implemented a state-of-the-art segmentation algorithm named ‘PanSegNet’ \cite{zhang2025large}. For this training, five radiologists, each representing one of the five centers, manually segmented the pancreas on axial $385$ T1w and 382 T2w scans from $499$ adult patients across the centers. Before the annotation process, the radiologists established a protocol to standardize their approach, and all used the same software, ITK-SNAP \cite{yushkevich2016itk} for these annotations \cite{tustison2010n4itk}. In the second round, a total of $341$ T1W and $364$ T2-weighted (T2W) scans from NU and two additional centers, IU and EMC, were automatically segmented using PanSegNet. Two abdominal radiologists then reviewed and corrected the segmentations to establish the ground-truth segmentations for the second round segmentations. 

\subsection{{Pre-Processing and Radiomic Feature Map Extraction}}
All imaging data and their corresponding annotations were adjusted to a uniform isotropic resolution of $1 \times 1 \times 1$ mm\(^3\)
in the $X$, $Y$, and $Z$ axes. For each region of interest (ROI), 13 GLCM (Gray Level Co-occurrence Matrix) features \cite{haralick1973textural} were derived on a voxel-wise basis, employing a local $3 \times 3 \times 3$ sliding window to capture intensity co-occurrences across all spatial dimensions. These GLCM features, frequently utilized in radiomic studies, \cite{prasanna2017radiomic}, produced 13 distinct voxel-wise GLCM radiomic feature maps for each ROI, labeled $I_1$ through $I_{13}$.

\subsection{{GrRAiL Development}}

The algorithm was implemented in Python to compute $\textbf{Q}$ for each $I_1, \ldots, I_{13}$. The features were specifically selected to capture the 15 global attributes of the graph, per radiomic expression map, with the number and type of features kept consistent across all experiments. The extracted graph-theoretic measurements ($15 \times 13 = 195$) are concatenated to form the GrRAiL descriptor, then evaluated for different diagnostic tasks using machine learning classifiers. Finally, to evaluate the impact of different bin sizes on the extracted features and the model's overall performance, we conducted a parameter sensitivity analysis by varying the number of discrete intensity levels (bins) into which continuous voxel intensities are divided, using values of $4$, $16$, and $64$.

To identify the optimal number of clusters to spatially segment radiomic expression maps using a GMM, we used the BIC in different cluster combinations, ranging from $1$ to $12$. Empirical analysis consistently identified that ${{u}}=5$ clusters yielded the highest predictive performance across different use-cases, while maintaining reduced computational load and time requirements. 

\subsection{Experimental Setup}

For all analyses, the classification models were trained and optimized using 5-fold cross-validation on the training cohort, and their performance was subsequently evaluated on a hold-out test cohort. Specifically, we performed feature selection followed by random forest (RF) classification to distinguish the different pathologies and obtain performance metrics. Recursive feature elimination \cite{guyon2002gene} was applied to the GrRAiL descriptor to obtain uncorrelated features. These features were then employed in an RF classifier on the training set to distinguish the different pathologies. The top features resulting from the training model were then used on the test set for the classification task. Performance metrics, including the area under the receiver operating characteristic curve (AUC), cross-validation (CV) accuracy, and test accuracy, were used to assess the diagnostic performance of GrRAiL.
To evaluate the contributions of different features that comprise GrRAiL, we performed SHapley Additive exPlanations (SHAP) analysis \cite{lundberg2017unified} on the graph features to identify the top 10 descriptors that contributed most significantly to the model's performance. We conducted statistical analysis using the Mann-Whitney U test on the SHAP-identified features to provide insights into the discriminatory ability of different features within the GrRAiL  descriptor, across different use-cases. Additionally, statistical differences in model performance metrics (test accuracy) between GrRAiL and comparative methods were evaluated using a two-tailed z-test, to determine statistical significance of observed performance differences.

\subsection{Ablation Studies}

\subsubsection{Graph Neural Networks for Adjacency Matrices Analysis}

A weighted region adjacency graph (RAG) was created from $K$ clusters for each radiomic feature map, with edge weights computed using the EMD. From the constructed graph $G = (V, E)$, the adjacency matrix $A = [A_{ij}]$ was computed, resulting in a total of $13$ adjacency matrices ($A_1, \ldots, A_{13}$) derived from $I_1, \ldots, I_{13}$, respectively. The resulting adjacency matrices were evaluated using Graph Neural Network algorithms. 

We conducted multiple experiments to identify the optimal approach for processing the generated adjacency matrices from the constructed graphs using several state-of-the-art (SOTA) Graph Neural Network (GNN) algorithms. Specifically, we selected four GNNs for graph-level classification: GraphSAGE \cite{Hamilton2017GraphSAGE}, Graph Isomorphism Network (GIN) \cite{Xu2018GIN}, Graph Attention Network (GAT) \cite{Velickovic2018GAT}, and Graph Convolutional Network with Jumping Knowledge (GCN-JK) \cite{Xu2018GCNJK}. To ensure a fair comparison, we maintained consistent training hyperparameters and loss functions across all models. GraphSAGE (Graph Sample and Aggregation) generates node embeddings by sampling and aggregating features from a node's local neighborhood, enabling generalization to unseen nodes and dynamic graphs. In our study, we used initial node features derived from several graph properties, such as node degrees, closeness centrality, clustering coefficients, and average neighbor degree. The GraphSAGE model then used these embeddings for classification.
The GIN, known for its strong discriminative power, closely mimics the Weisfeiler-Lehman graph isomorphism test. The GAT employs attention mechanisms to assign varying importance to neighboring nodes, enabling it to focus on the most relevant parts of the graph. GCN-JK incorporates a Jumping Knowledge mechanism, which aggregates information from various layers of the network. 
For GraphSAGE and GCN-JK, the $13$ adjacency matrices were aggregated and directly input into the neural network models to derive performance metrics that were similarly computed for GrRAiL classification.

\subsubsection {GLCM Radiomics Analysis}

To assess GrRAiL's performance, a comparative assessment that employs GLCM-based features in a voxel-wise manner was conducted. Similar to the GrRAiL approach, recursive feature elimination was applied to the voxel-wise radiomic descriptors extracted from individual radiomic feature maps ($I_1, \ldots, I_{13}$) to obtain uncorrelated features. Then, for each of the 13 voxel-wise GLCM features, we calculated five statistical descriptors: mean, median, standard deviation, kurtosis, and skewness \cite{gillies2016radiomics,VanGriethuysen2017}. These descriptors capture the distribution of radiomic features across the ROI, resulting in a total of 65 voxel-wise descriptors. These features were then employed in an RF classifier on the training set to distinguish the different pathologies. The top features resulting from the training model were then used on the test set for the classification task. 

\subsubsection{Intensity Graph Analysis}

For this experiment, we performed clustering directly on the tumor region using a GMM rather than generating radiomic expression maps first, then performing clustering. BIC was used to determine the optimal number of clusters, ranging between $1$ and $5$. Nodes in the graph represent the centroids of each cluster, while edges were computed using the EMD, resulting in a graph that encapsulates diverse patterns of heterogeneity from the intensity clusters \cite{Battalapalli2024SIGL}. From the constructed weighted graph, we extracted $15$ graph theory-based features to evaluate the graph architecture and quantify lesion complexity.

\section{RESULTS}

\subsection{Experiment 1: Distinguishing Tumor Recurrence from Pseudoprogression in Glioblastoma}

\subsubsection{GrRAiL Analysis}
Figure 3 illustrates qualitative 3D GLCM entropy feature maps alongside their corresponding GrRAiL-generated graphs for glioblastoma patients with lesions exhibiting TuR  (a, d) and those exhibiting PsP (g, j). The GrRAiL features appear to be capturing the distinctive spatial patterns from the corresponding radiomic textural maps across PsP (i, l) and TuR (c, f). For instance, lesions characterized as PsP largely exhibited a markedly lower number of nodes and edges in the graph, suggesting greater homogeneity within the lesion microenvironment. Conversely, the larger $n$ of nodes found in the graphs corresponding to patients with TuR reflected higher heterogeneity within the lesion microenvironment. These trends were consistently observed across both training and test cohorts of patients with PsP and TuR analyzed using GrRAiL. 

\begin{figure*}[t!]
\centering
\includegraphics[height=11cm, width=0.95\textwidth]{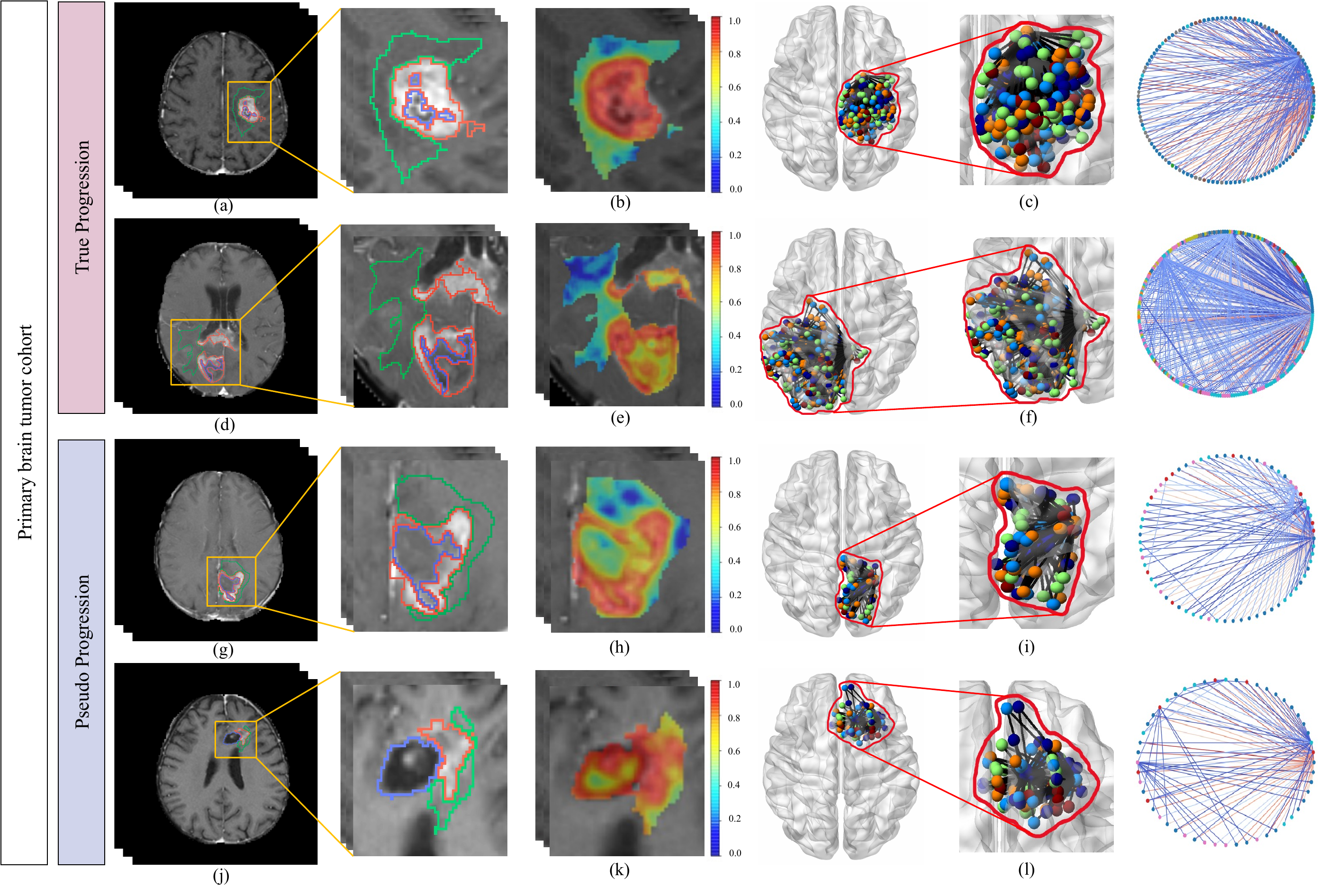}
\caption{\label{fig:res1}(a, d) and (g,j) demonstrate Gd-T1w MRI scans for TuR and PsP glioblastoma studies, respectively. (b, e) and (h,k) show the GLCM entropy expression maps, while (c,f) and (i,l) are the GrRAiL-generated graphs from TuR and PsP subjects, respectively. Note the differences in complexity of GrRAiL graphs in the last column across TuR and PsP cases.}
\end{figure*}

\begin{figure*}[t!]
\centering
\includegraphics[height=13cm, width=1\textwidth]{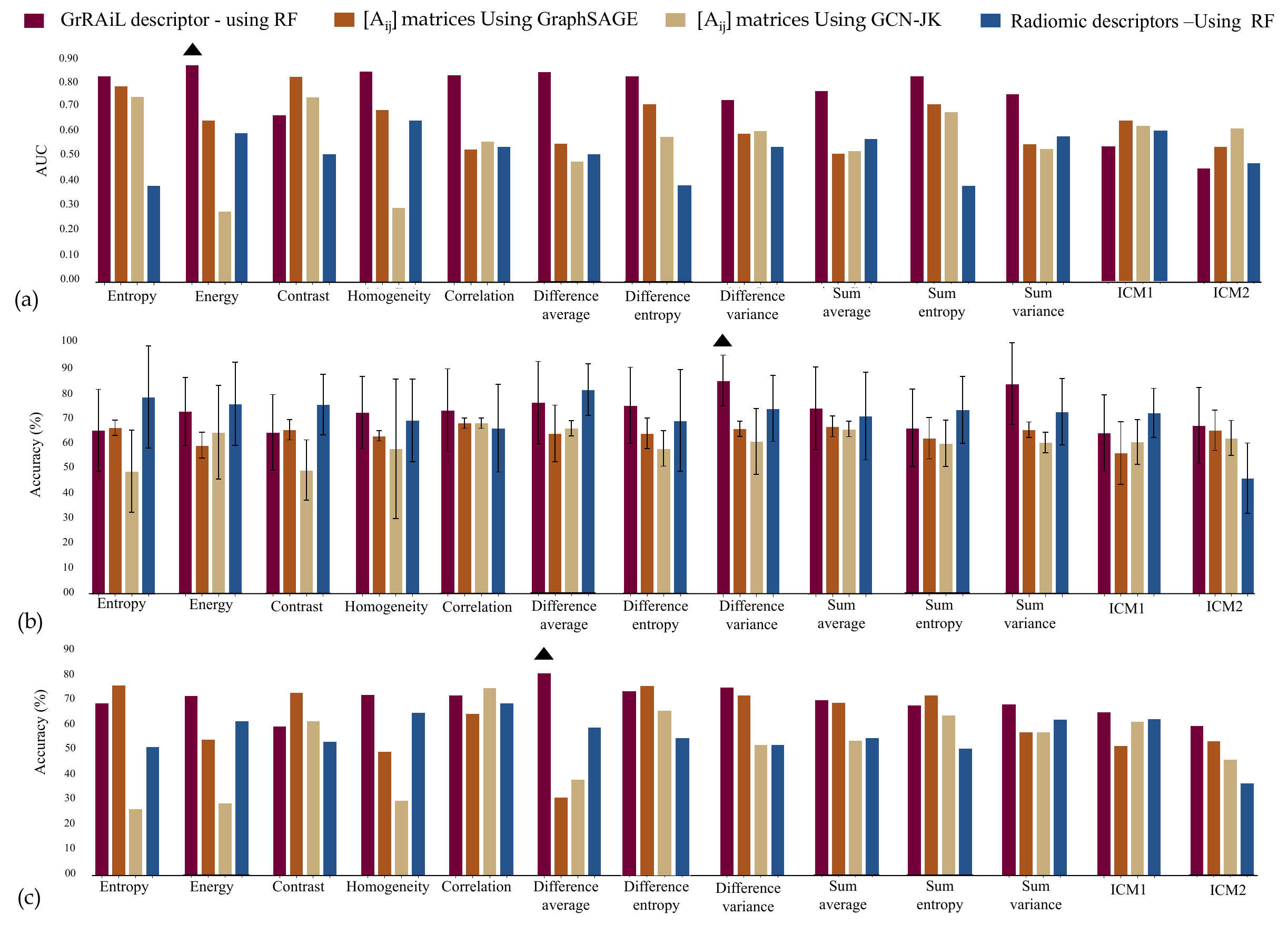}
\caption{\label{fig:res1}Classification performance metrics for GrRAiL, GraphSAGE, GCN-JK, and radiomic methods across 13 GLCM features in the context of glioblastoma tumors for the classification of TuR from PsP. The bar plots correspond to (a) AUC values, (b) cross-validation accuracy values on the training data, as well as(c)  test accuracy values on unseen data from a different institution.}
\end{figure*}

Bar plots corresponding to the performance metrics (AUC, CV accuracy, and test accuracy) for each radiomic feature map are shown in Figure 4. Among the 13 radiomic maps, the GrRAiL feature from the energy map achieved the highest AUC value ($0.86$), the difference variance map yielded the highest CV accuracy ($83\% \pm 10\%$), and the difference average map yielded the highest test accuracy ($78\%$), on the test set. The top-performing metrics in Figure 4 are highlighted with a delta above the corresponding bar plot. The performance metrics of GrRAiL are shown in Figures 5a and 5b. The AUC value was $0.85$, the CV accuracy was $89\% \pm 7\%$, and the accuracy on the test set was $78\%$. The SHAP summary plot in Figure 5c highlights the top 10 GrRAiL features that significantly contributed to the classification accuracies for distinguishing PsP from TuR. In Figure 5((d)-(h)), we highlighted the features with p-value $\leq 0.01$, indicated with a single asterisk above the box plot. These features include the assortativity graph feature from the ICM2 expression map, modularity from the energy expression map, network entropy from the ICM2 expression map, average path length from the contrast expression map, clustering coefficient from the homogeneity expression map, and small-worldness from the homogeneity expression map. Comparison of GrRAiL test-set accuracy metric with other comparative approaches using z-statistics, and the corresponding p-values are detailed in Supplementary Table 7.

\subsubsection{Ablation Studies}

 \textit{Assessment of graph-derived adjacency matrices:} The adjacency matrices obtained for each graph from individual radiomic feature maps through GraphSAGE and GCN-JK neural networks using more than 2 convolutional layers (e.g., $5, 7$) did not yield significant improvement; instead, the accuracy values decreased. Thus, these two convolutional layers were used for further analysis (Figure 4). GraphSAGE achieved an AUC of $0.80$ using the contrast map. The entropy map yielded a test accuracy of $74\%$, and the contrast map yielded the highest CV accuracy of $69.3\% \pm 11\%$. GCN-JK achieved an AUC of $0.73$ using the entropy map. The correlation map yielded a CV accuracy of $66\% \pm 2\%$, and a test accuracy of $73\%$, as shown in Figure 4. Furthermore, when the adjacency matrices from all 13 radiomic feature maps were combined and fed into GraphSAGE and GCN-JK, GraphSAGE achieved a CV accuracy of $59\% \pm 3\%$, its performance on the test set was 64\% and an AUC of $0.59$. The GCN-JK model yielded a CV accuracy of $62\% \pm 0.4\%$, a test accuracy of $66\%$, and an AUC of $0.61$, suggesting that the model may be limited in its ability to capture discriminatory features across PsP and TuR (Figures 5a, 5b). Results, when applying the GAT and GIN schemes, are provided in supplementary Table 2. Both GAT and GIN yielded poor performance in distinguishing PsP from TuR, compared to GrRAiL.

\textit{Assessment of radiomic features:} Performance metrics (AUC, CV accuracy, and test accuracy) for each radiomic feature map are depicted using bar plots in Figure 4. Among the 13 radiomic maps, the radiomic descriptors from the homogeneity map achieved the highest AUC value ($0.63$), while the contrast map yielded the highest CV accuracy ($79\% \pm 10\%$) and correlation map achieved a test accuracy of ($68\%$). Additionally, when the radiomic descriptors from all 13 radiomic feature maps were combined into the classification model, the performance metrics showed an increase in the AUC and test accuracy values, as shown in Figures 5a and 5b. The AUC value was $0.79$, CV accuracy was $72\% \pm 13\%$, and the test accuracy was $69\%$. Overall, combining features from all 13 radiomic maps significantly enhanced the radiomic analysis performance but led to decreased performance for the adjacency matrices. When comparing individual analyses, GrRAiL outperformed both adjacency matrices and radiomic descriptors in the classification task.

\textit{Assessment of Intensity Graph Analysis:} The intensity graph analysis yielded a cross-validation accuracy of $69\% \pm 11\%$, a test accuracy of $59\%$, and an AUC value of $0.56$, respectively. Both GrRAiL and the radiomic analysis was found to outperform the intensity graph analysis, with statistically significant improvements in accuracy (p-value = 0.04, z-score = 2.019) (Supplementary Table 7). The performance metric values are provided in supplementary Table 1.

\begin{figure*}[t!]
\centering
\includegraphics[height=11.5cm, width=1\textwidth]{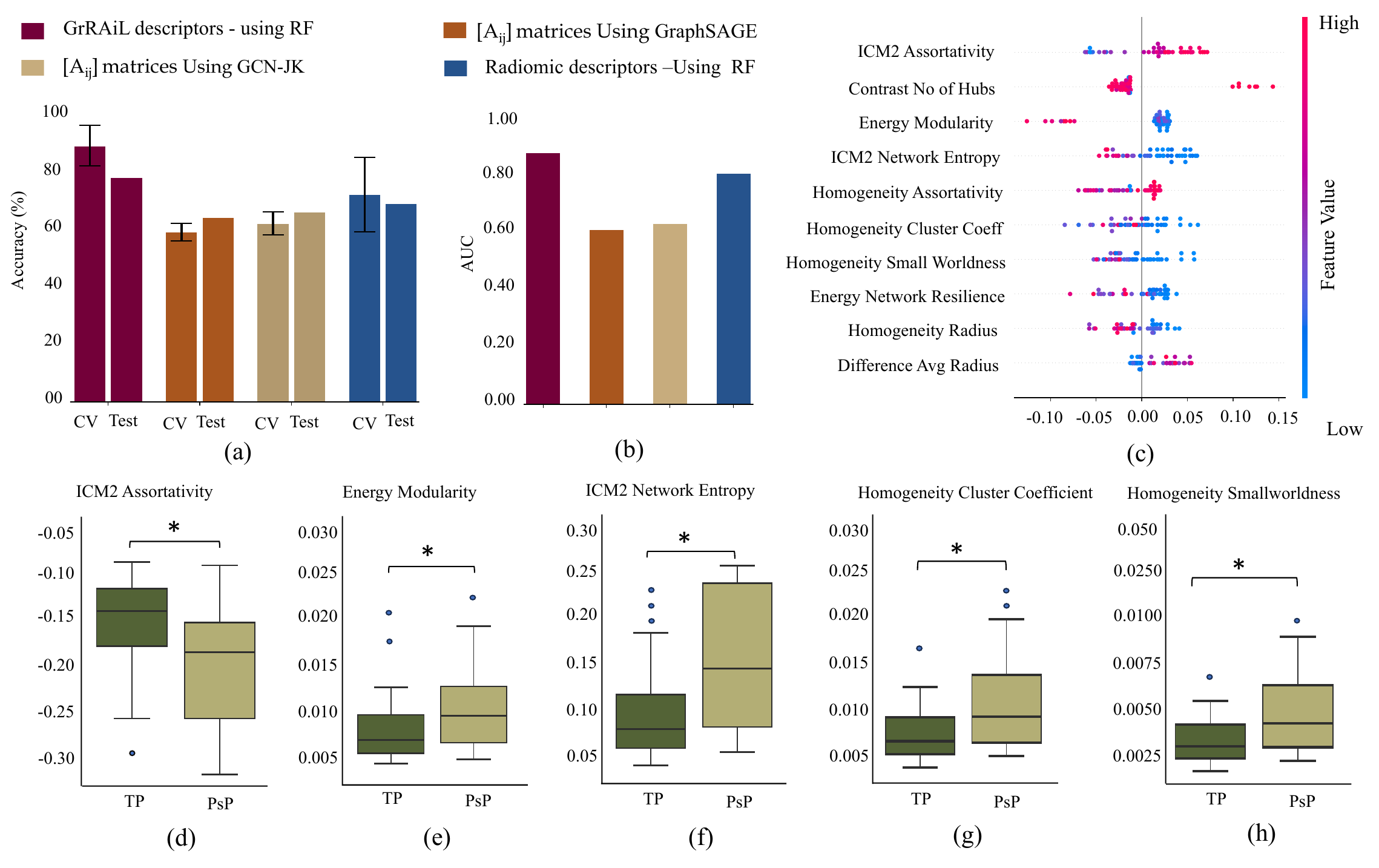}
\caption{\label{fig:res1}  Bar plots with (a) CV and test accuracy, and (b) AUC values for graph features,  GrRAiL, GraphSAGE, GCN-JK, and Radiomics features, (c) SHAP summary plot for the top 10 features contributing to the GrRAiL classification accuracy in distinguishing TuR from PsP in glioblastoma tumors. The remaining box plots are based on the statistical significance with $p \leq 0.01$ and include (d) ICM2 Assortativity, (e) Energy Modularity, (f) ICM2 Network Entropy, (g) Homogeneity Cluster coefficient, and (h) Homogeneity small worldness. Each feature demonstrated significant differences across TuR and PsP, highlighting the efficacy of GrRAiL features in differentiating true tumor progression from PsP.}
\end{figure*}

\begin{figure*}[t!]
\centering
\includegraphics[height=12cm, width=0.95\textwidth]{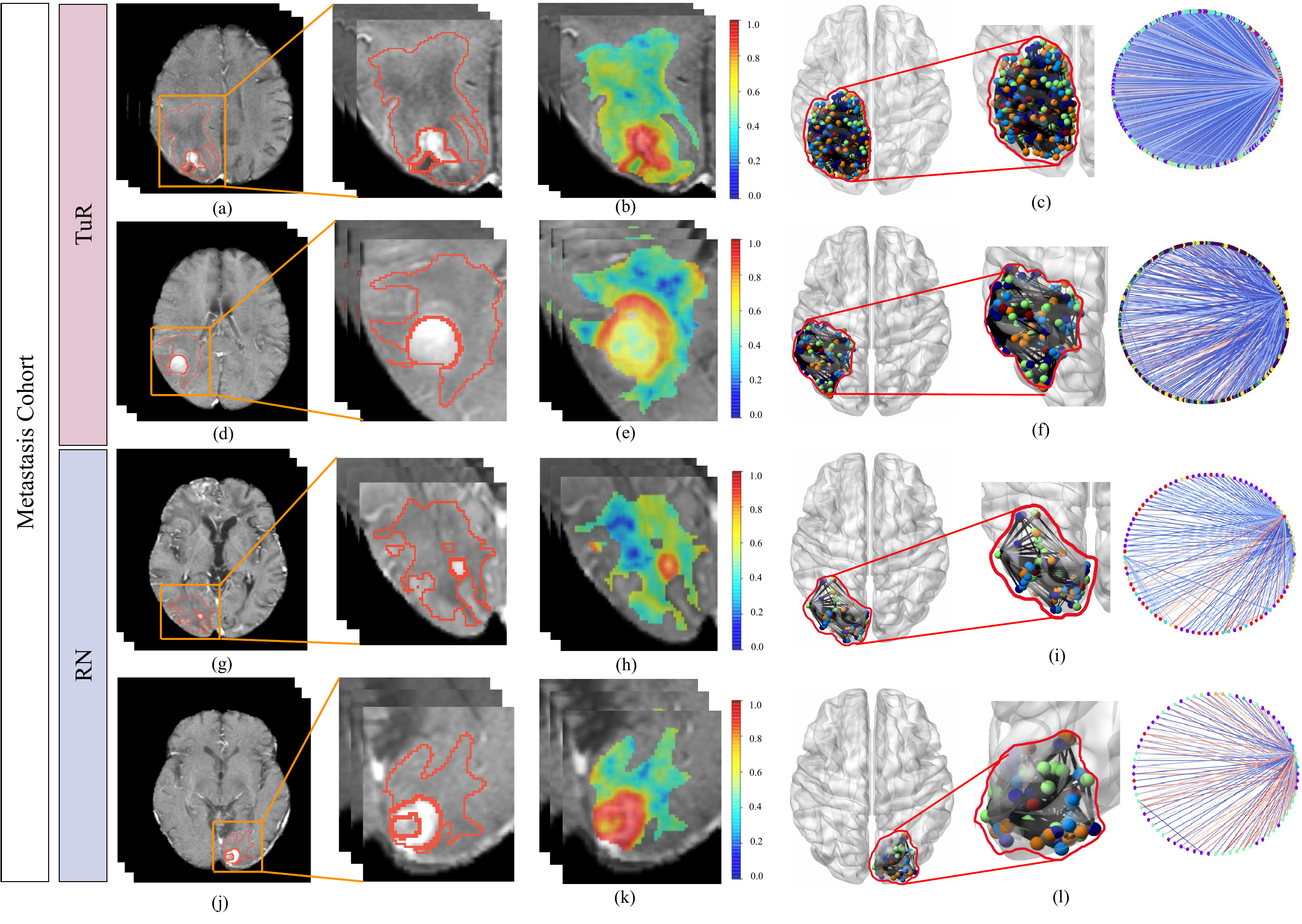}
\caption{\label{fig:res1}(a) and (d) demonstrate Gd-T1w MRI scans for TuR and RN, respectively, for two metastatic brain tumor studies. (b) and (e) highlight the corresponding GLCM entropy expression maps, while the corresponding GrRAiL generated graphs for TuR and RN are shown in (c) and (f), respectively. Note the differences in complexity of GrRAiL graphs in the last column across TuR and PsP cases in the metastatic brain tumor cohort.}
\end{figure*}

\begin{figure*}[t!]
\centering
\includegraphics[height=13.3cm, width=1.0\textwidth]{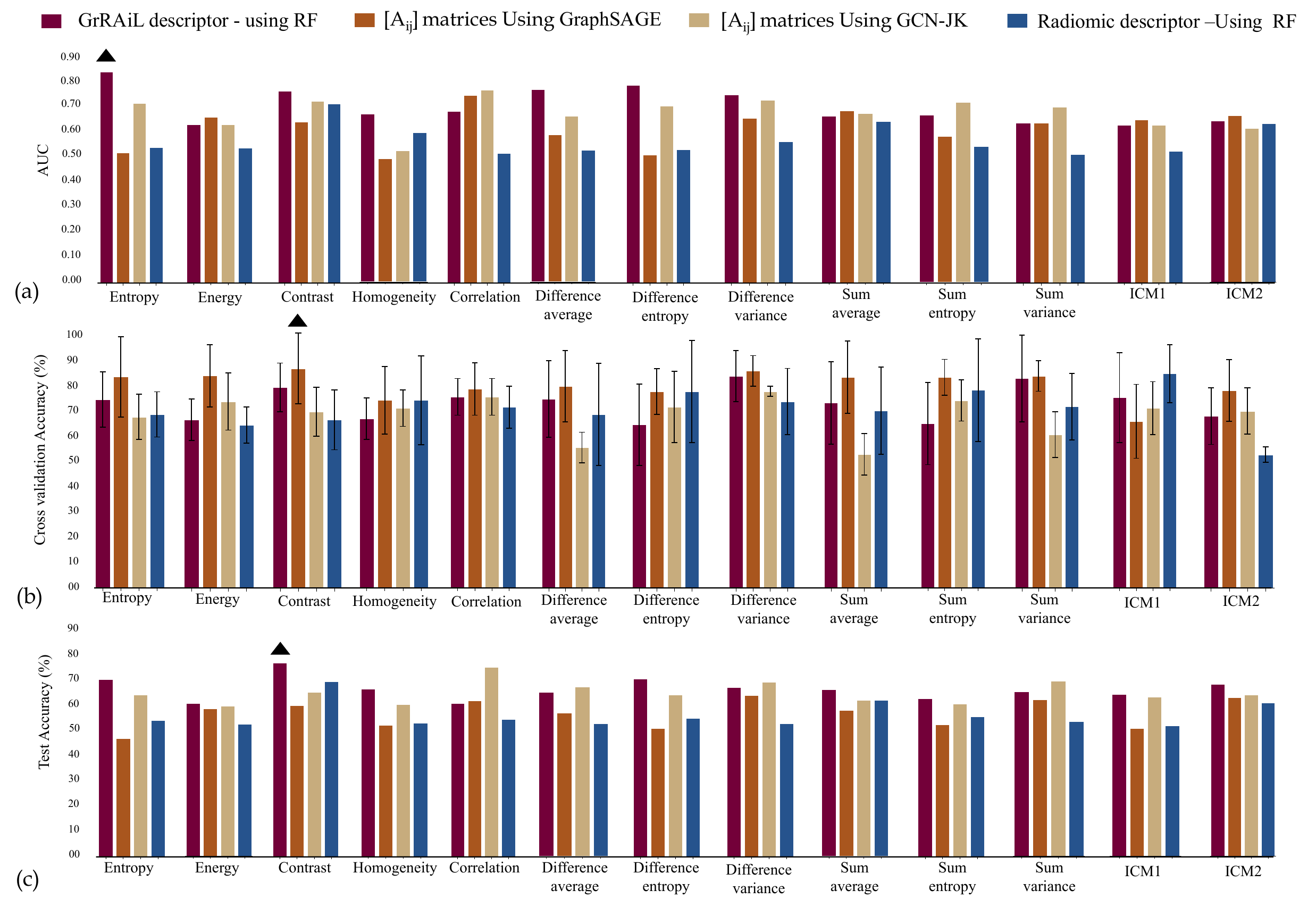}
\caption{\label{fig:res1}Classification performance metrics for GrRAiL, GraphSAGE, GCN-JK, and radiomic methods across 13 GLCM features in the context of metastasis brain tumors for the classification of TuR from RN.  Bar plots of (a) AUC values, (b) cross-validation accuracy values on the training data, and (c) test accuracy values reflecting the model's ability to generalize to unseen data.}
\end{figure*}

\begin{figure*}[t!]
\centering
\includegraphics[height=11cm, width=1\textwidth]{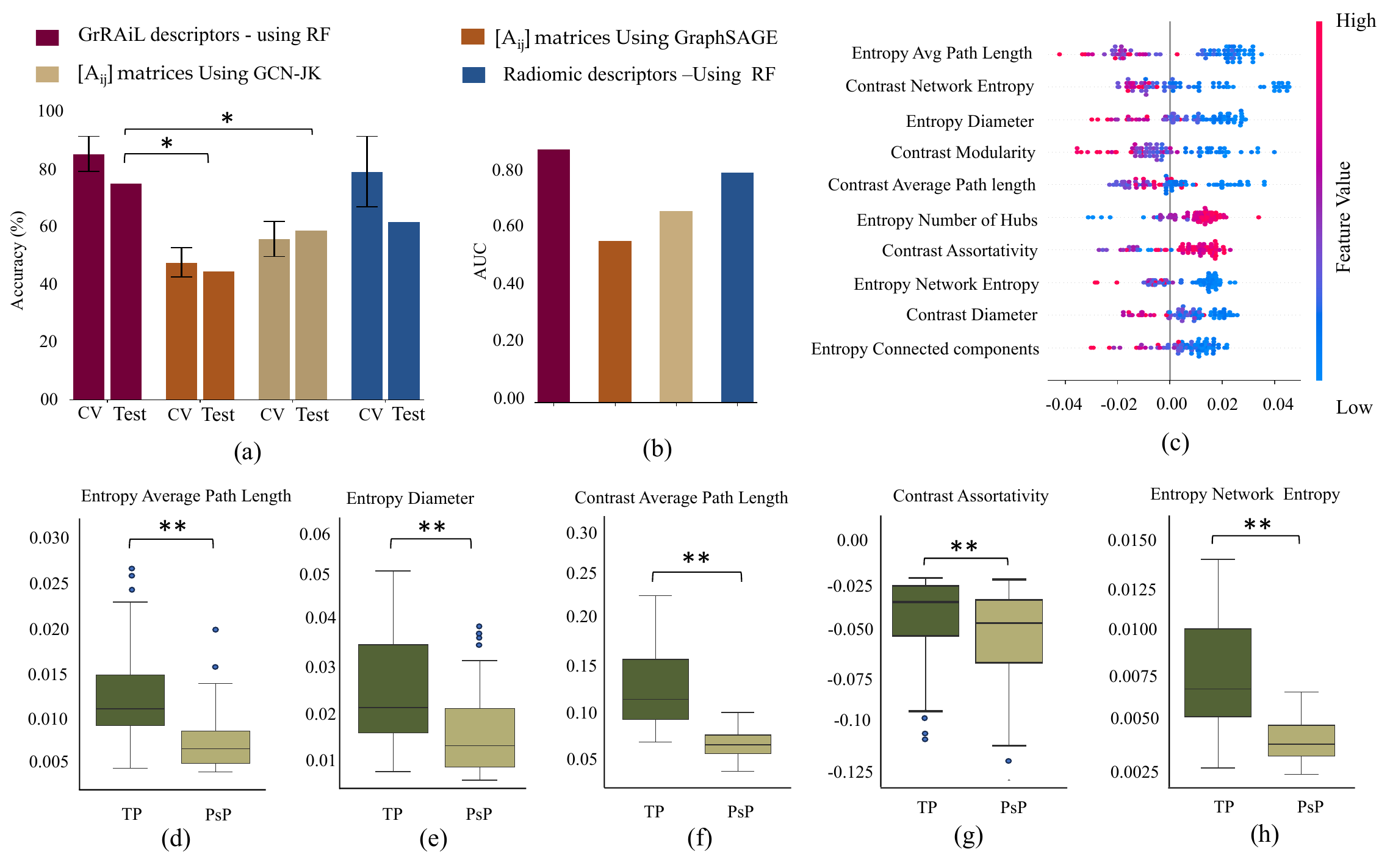}
\caption{\label{fig:res1} (a) Bar plots with cross validation accuracy and test accuracy for GrRAiL, GraphSAGE, GCN-JK and Radiomics features from metastasis cohort (b) Bar plots with AUC values for graph features, GraphSAGE, GCN-JK and Radiomics features from metastasis cohort (c) SHAP summary plot for the top 10 features contributing to the GrRAiL classification accuracy in distinguishing TuR from RN in metastasis brain tumors. The remaining box plots are based on the statistical significance with p $\leq$ 0.0001 and include (d) Entropy average path length, (e) Entropy diameter, (f) Contrast average path length, (g) Contrast assortativity and (h) Entropy network entropy. Each feature demonstrates significant differences between TuR and PsP, highlighting the efficacy of graph-based features in differentiating true tumor progression from PsP.}
\end{figure*}

\begin{figure*}[t!]
\centering
\includegraphics[height=11.5cm, width=1.0\textwidth]{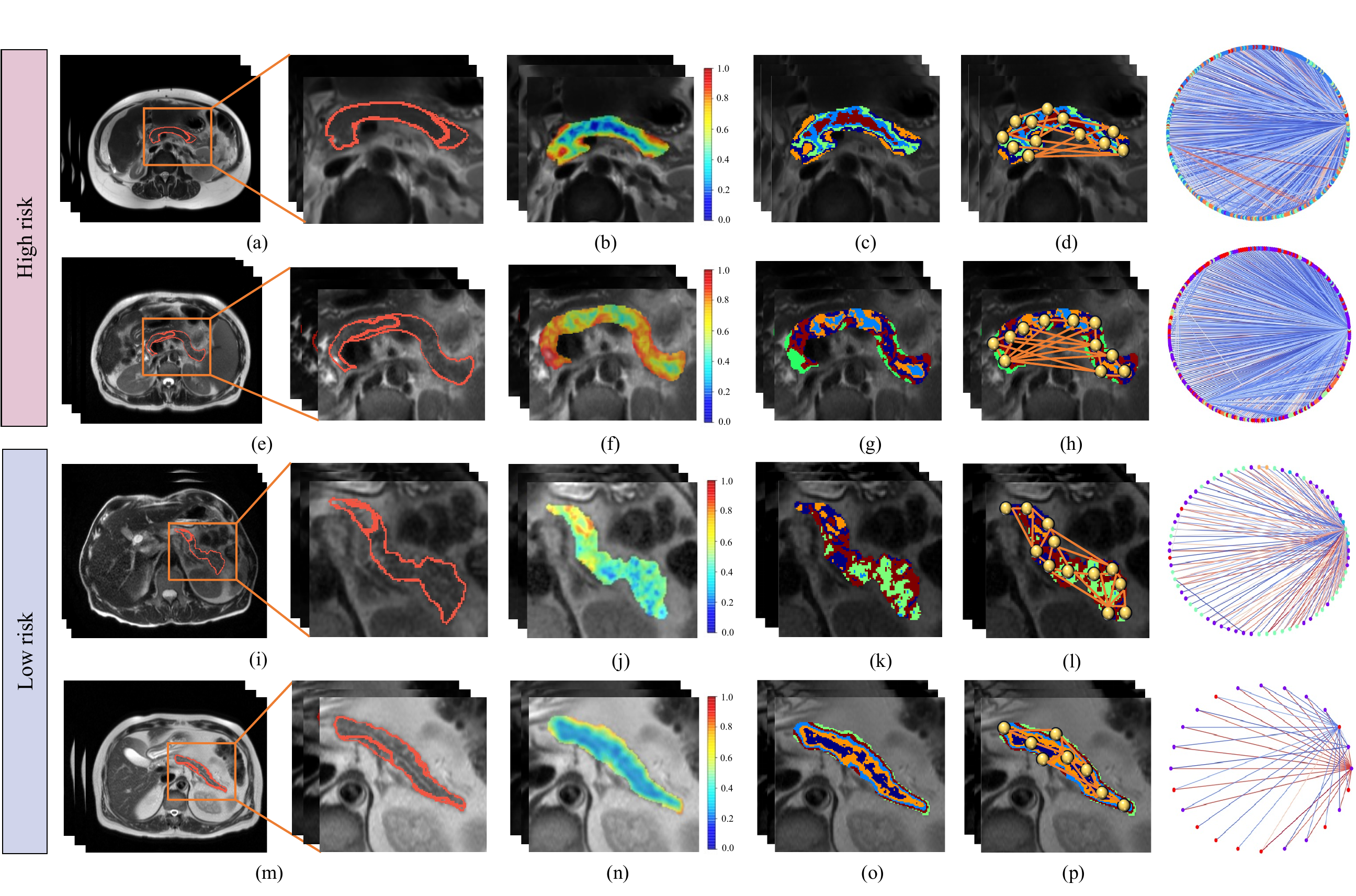}
\caption{\label{fig:res1}(a, e) and (i, m) demonstrate T2w MRI scans for high-risk and low-risk IPMN studies, respectively. (b, f) and (j, n) show the GLCM entropy expression maps, while (c, g) and (k, e) are the corresponding cluster maps for high-risk and low-risk subjects, respectively. Finally, (d, h) and (i, p) are the GrRAiL-generated graphs from high-risk and low-risk subjects, respectively. Note the differences in complexity of GrRAiL graphs in the last column across high versus low-risk IPMN cases.}
\end{figure*}

\begin{figure*}[t!]
\centering
\includegraphics[height=13.3cm, width=1.0\textwidth]{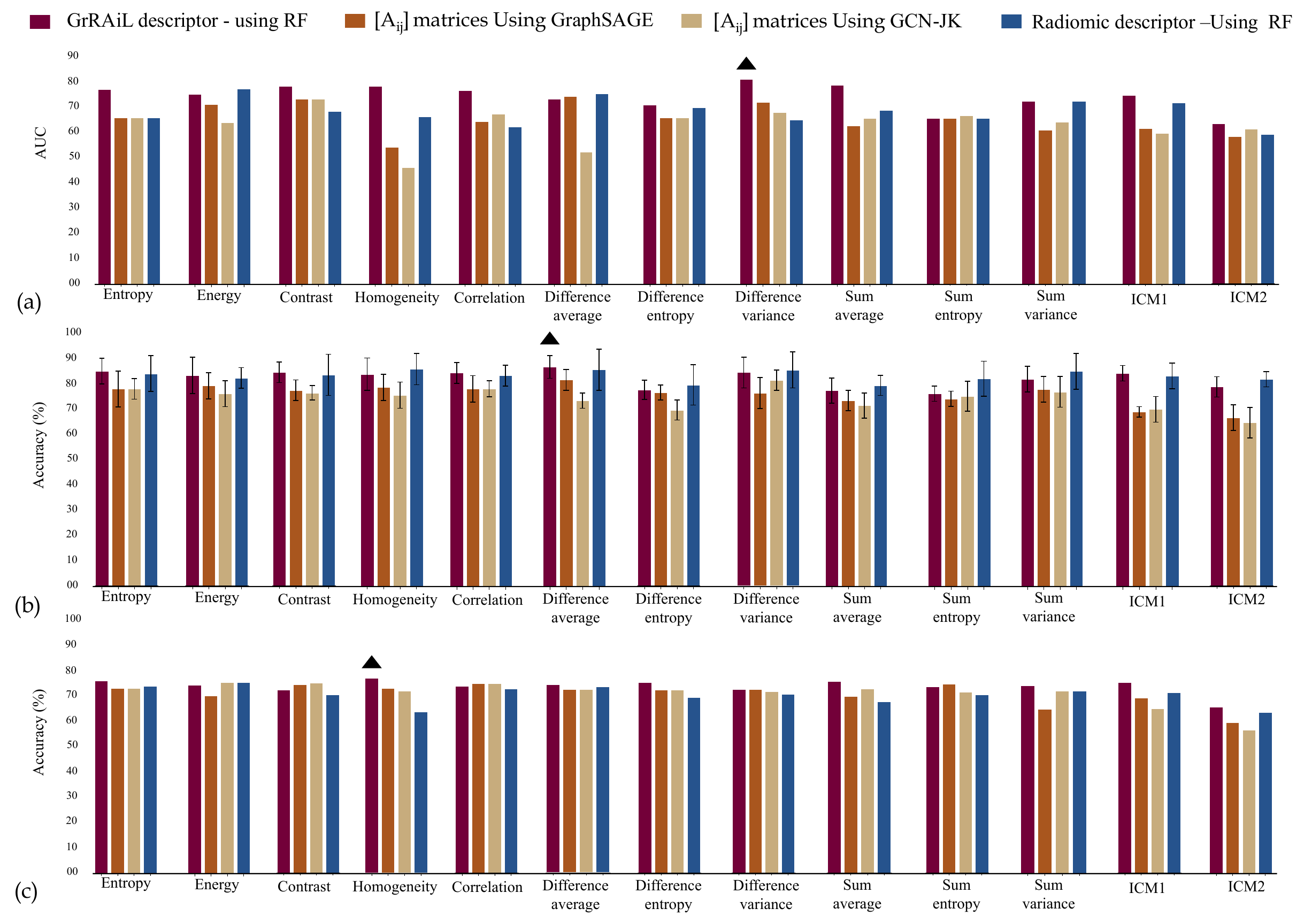}
\caption{\label{fig:res1}Classification performance metrics for GrRAiL, GraphSAGE, GCN-JK, and radiomic methods across 13 GLCM features in the context of IPMN for the classification of high risk from no + low risk lesions. Bar plots of (a) AUC values,  (b) cross-validation accuracy values on the training data, and (c) test accuracy values reflecting the model's ability to generalize to unseen data.}
\end{figure*}

\begin{figure*}[t!]
\centering
\includegraphics[height=11.5cm, width=1\textwidth]{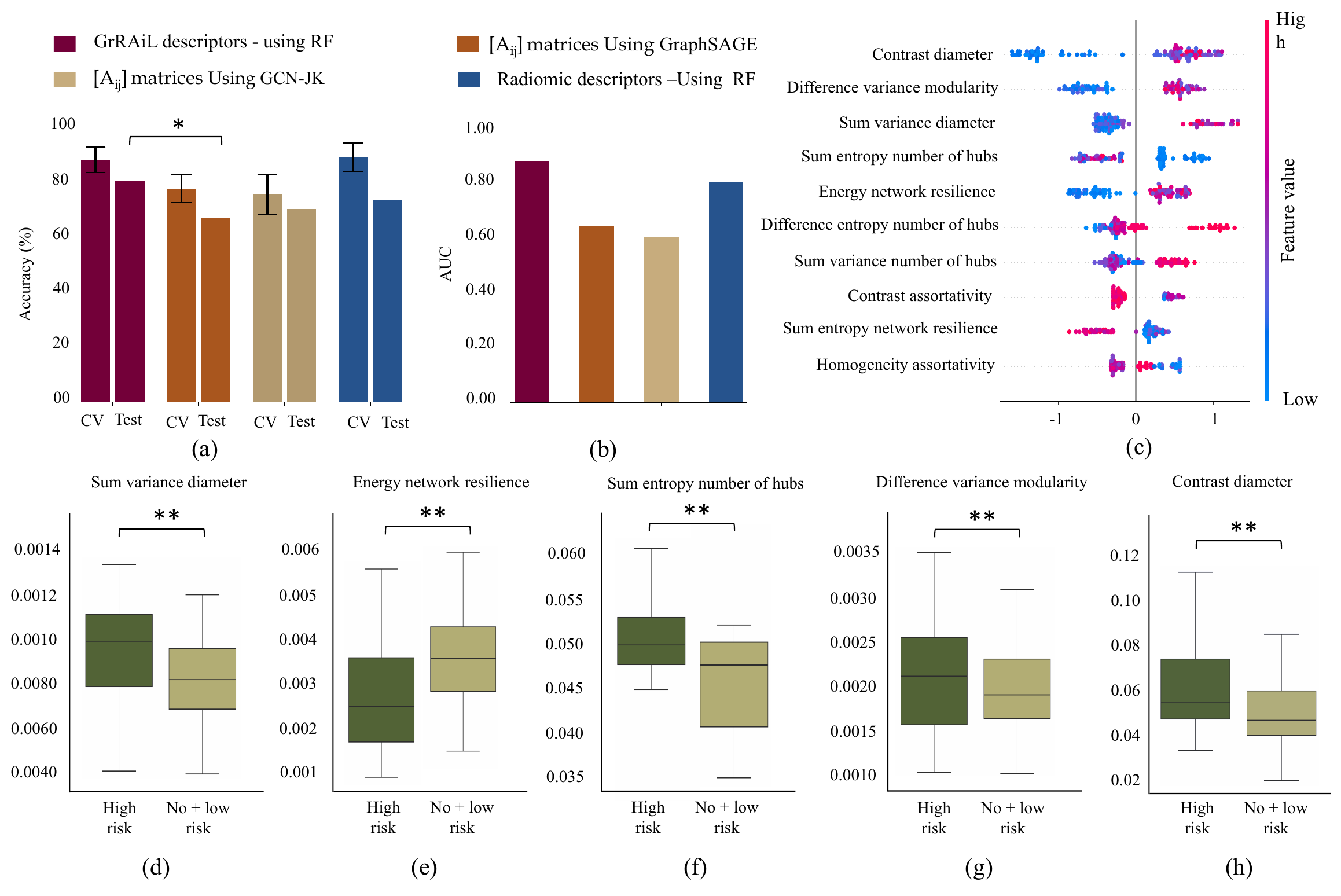}
\caption{\label{fig:res1} (a) Bar plots with cross validation accuracy and test accuracy for GrRAiL, GraphSAGE, GCN-JK and Radiomics features from IPMN cohort (b) Bar plots with AUC values for graph features, GraphSAGE, GCN-JK and Radiomics features from IPMN cohort (c) SHAP summary plot for the top 10 features contributing to the GrRAiL classification accuracy in distinguishing high risk from No + low risk lesions in IPMN patients. The remaining box plots are based on the statistical significance with p $\leq$ 0.001 and include (d) sum variance diameter, (e) Energy network resilience, (f) Sum entropy number of hubs, (g) difference variance modularity, and (h) Contrast diameter. Each feature demonstrates significant differences between high-risk and no + low-risk, highlighting the efficacy of GrRAiL features.}
\end{figure*}

\subsection{Experiment 2: Distinguishing Tumor Recurrence from Radiation Necrosis in Metastatic Brain Tumors}

\subsubsection{GrRAiL Analysis}
Figure 6 illustrates qualitative 3D GLCM entropy feature maps alongside their corresponding GrRAiL-generated graphs for patients with brain metastases. Similar to our findings in glioblastoma tumors, GrRAiL effectively captured unique spatial patterns from the radiomic texture maps across RN and TuR in metastatic brain tumors. Lesions classified as RN typically exhibited fewer nodes and edges in their graphs, indicating a higher degree of homogeneity within the lesion microenvironment. In contrast, TuR graphs displayed a greater number of nodes and edges, reflecting increased heterogeneity in lesions that corresponded to tumor progression. These observations were consistent across both training and test cohorts analyzed using GrRAiL. Performance metrics (AUC, CV accuracy, and test accuracy) for each radiomic feature map are depicted using bar plots in Figure 7. Among the 13 radiomic maps, the GrRAiL features from the entropy map achieved the highest AUC ($0.81$), and the highest CV accuracy ($75\% \pm 11\%$), and the contrast map yielded the highest test accuracy ($78\%$). The top-performing metrics in Figure 7 are highlighted with an delta above the corresponding bar plot. The performance metrics of GrRAiL descriptor are shown in Figures 8a and 8b. We obtained an AUC of $0.86$, CV accuracy of $84\% \pm 6\%$, and test accuracy of $74\%$ in distinguishing RN from TuR in the metastatic brain tumor cohort.  The SHAP summary plot in Figure 8c highlights the top 10 GrRAiL features that significantly contributed to classification accuracy with a focus on features that yielded a p-value of $\leq 0.001$, indicated by double asterisks on the box plots.
Statistical comparisons across GrRAiL and other comparative approaches using z-statistics and the corresponding p-values, are provided in Supplementary Table 7.

\subsubsection{Ablation Studies}

\textit{Assessment of graph-derived adjacency matrices:} GraphSAGE achieved an AUC of $0.71$ using the correlation map. The difference variance map resulted in a test accuracy of $62\%$, and the contrast map achieved the highest CV accuracy at $84\% \pm 16\%$. GCN-JK attained an AUC of $0.73$ with the correlation map, a CV accuracy of $77\% \pm 8\%$ using the sum variance map, and a test accuracy of $73\%$ with the correlation map. However, when adjacency matrices from all 13 radiomic maps were combined and input into GraphSAGE and GCN-JK, the performance metrics declined. GraphSAGE's CV accuracy was $47\% \pm 5$, its test accuracy dropped to 44\% with an AUC of $0.55$. GCN-JK gave a CV accuracy of $55\% \pm 7\%$ and a test accuracy of $58\%$, and an AUC of $0.65$, indicating relatively poor performance in distinguishing TuR from RN in metastatic brain tumors (Figure 8). Results using the GAT and GIN schemes are provided in supplementary Table 4. The statistical comparison of test accuracies revealed that GrRAiL significantly outperformed GraphSAGE (p-value = 0.0002, z-score = 3.786) and GCN-JK (p-value = 0.04, z-score = 2.05) (Supplementary Table 7).

\textit{Assessment of radiomic features:} Performance metrics for each radiomic feature map are shown in Figure 7. Among the 13 radiomic maps, the contrast map's radiomic descriptors achieved the highest AUC of $0.69$, the homogeneity map yielded the highest CV accuracy at $72\% \pm 7\%$, and the contrast map achieved a test accuracy of $66\%$. When we combined the radiomic descriptors from all 13 maps into the classification model, there was a slight increase in AUC and test accuracy, as shown in Figure 8. The AUC improved to $0.79$, CV accuracy was $78\% \pm 12\%$, and test accuracy was $61\%$. However, the overall performance remained lower compared to the GrRAiL features.

\textit{Assessment of intensity graph analysis:} The intensity graph analysis yielded a cross-validation accuracy of $73\% \pm 11\%$, a test accuracy of $55\%$, and AUC value of $0.55$. These performance metrics indicate that both GrRAiL as well as the radiomic analysis outperformed the intensity graph analysis.  Besides, there was a statistically significant difference between the test accuracies of the GrRAiL approach and the intensity graph approach (p-value = 0.017, z-score = 2.37) (Supplementary Table 7). The performance metric values are provided in supplementary Table 3.

\subsection{Experiment 3: Distinguishing No-Risk + Low-Risk vs. High-Risk Pancreatic IPMN lesions}

\subsubsection{GrRAiL Analysis}
Figure 9 presents qualitative 3D GLCM entropy feature maps and their corresponding GrRAiL-generated graphs for patients categorized as having no-risk + low-risk versus high-risk IPMN lesions. 
Lesions in the no-risk + low-risk IPMNs category exhibited a lower number of nodes and edges in their graph representations, suggesting a more homogeneous tissue microenvironment.  In contrast, patients classified as high-risk displayed graphs with a significantly larger number of nodes and edges, reflecting increased lesion heterogeneity.

Performance metrics (AUC, CV accuracy, and test accuracy) for each radiomic feature map are shown in Figure 10. Among the 13 radiomic maps, the difference average map yielded the highest CV accuracy of $84\% \pm 4.3\%$, the homogeneity map achieved the highest test accuracy of $75\%$, and the difference variance maps showed an AUC of $0.79$. The top-performing metrics in Figure 10 are highlighted with an delta. The performance of GrRAiL is illustrated in Figure 11, where we obtained an AUC value of $0.83$, CV accuracy of $85\% \pm 4.5\%$, and test accuracy of $79\%$. Figure 11 presents the SHAP summary plot highlighting the top 10 GrRAiL features that contributed significantly to classification accuracy, with features demonstrating p-values $\leq 0.001$ (indicated by double asterisks above the box plots) in distinguishing no-risk + low-risk versus high-risk IPMN lesions. The z-statistics and the corresponding p-values towards statistical comparisons across GrRAiL test accuracy metric and other comparative approaches are detailed in Supplementary Table 7. 

\subsubsection{Ablation Studies}

\textit{Assessment of graph-derived adjacency matrices:} 
As presented in Figure 11, GraphSAGE achieved an AUC of $0.73$ using the contrast map, while the difference average map resulted in the highest CV accuracy of $80\% \pm 4\%$ and contrast map achieved a test accuracy of $73\%$. Similarly, GCN-JK achieved an AUC of $0.73$ using the contrast map, while the difference variance map resulted in the highest CV accuracy of $80\% \pm 4\%$ and the contrast map achieved a test accuracy of $72.2\%$. Interestingly, when adjacency matrices from all 13 radiomic maps were combined and input into GraphSAGE and GCN-JK, the performance declined, where GraphSAGE's CV accuracy dropped to $75.2\% \pm 5\%$ and its test accuracy dropped to $65\%$ with an AUC of $0.61$. GCN-JK showed similar trends, with a CV accuracy of $73.3\% \pm 7.4\%$, a test accuracy of $68\%$, and an AUC of $0.57$, demonstrating limited effectiveness when adjacency matrices from multiple radiomic maps were combined. Results using the GAT and GIN schemes are provided in
supplementary Table 6. GrRAiL was found to statistically outperform GraphSAGE for identifying high-risk from no-risk+low-risk IPMNs on the test-set (p-value = 0.02, z-score = 2.209) (Supplementary Table 7).

\textit{Assessment of radiomic features:} Performance metrics for individual radiomic feature maps are shown in Figure 10. Among the 13 radiomic maps, the energy map radiomic feature achieved the highest AUC of $0.75$, while the difference average map yielded the highest CV accuracy of $85\% \pm 8\%$, and the entropy map achieved a test accuracy of $72.14\%$. When radiomic features from all 13 maps were combined, classification performances improved slightly. As shown in Figure 11, the AUC increased to $0.76$, the CV accuracy to $86\% \pm 5\%$, and the test accuracy to $71\%$. However, GrRAiL still outperformed radiomic feature-based classification in every metric.

\textit{Assessment of intensity graph analysis:} The intensity graph analysis achieved a cross-validation accuracy of $83\% \pm 3.2\%$, a test accuracy of $61\%$, and an AUC of $0.52$. These performance metrics indicate that both GrRAiL and radiomic feature analysis significantly outperformed the intensity graph approach. Besides, there was a statistically significant difference between
the test accuracies of the GrRAiL approach and the intensity
graph approach (p-value = 0.005, z-score = 2.78) (Supplementary Table 7). The performance metric values are provided in supplementary Table 5. 

\section{DISCUSSION}

In this work, we presented a new descriptor called Graph-Radiomic Learning (GrRAiL), which leverages a combination of radiomic and graph-based attributes to quantify the intratumoral heterogeneity, towards distinguishing confounding pathologies from malignant lesions on MRI scans. Specifically, GrRAiL was employed to differentiate between tumor recurrence and treatment-induced radiation effects in glioblastoma (GB) and metastatic brain tumors, as well as to classify  Intraductal papillary mucinous neoplasms, the most common cystic tumors in pancreatic cancer, according to their risk level (no+low versus high-risk). This was achieved by using radiomic expression maps that capture local heterogeneity within the tumor region, and then connecting these heterogeneity patterns through a graph, followed by extracting graph-theoretic measurements that construct GrRAiL, to classify the different pathologies addressed in this work. Our results showed that GrRAiL outperformed the state-of-the-art approaches, e.g., the graph-derived adjacency matrices, in classifying the different pathologies across brain and pancreatic lesions. Notably, our results support the hypothesis that the global graph features that GrRAiL comprises may capture the spatial interactions of lesion `sub-populations', thus allowing for effectively representing heterogeneity within the lesion. Our findings demonstrated superior diagnostic performance compared to previous work in differentiating the different pathologies evaluated in this study, as detailed below.

\subsection{Distinguishing Tumor Recurrence versus Pseudoprogression in Glioblastoma}

 Our results showed that Assortativity (Figure 5d; $p=0.0049$)—which measures the tendency of graph nodes to connect to other nodes with similar attribute values \cite{Bullmore2009}—was higher in PsP than in TuR. In other words, clusters of voxels with similar GLCM ICM2 values in pseudo-progression lesions tended to form more tightly interconnected neighborhoods, which may reflect localized, treatment-related homogeneity. By contrast, the lower assortativity in true-progression graphs suggests that cluster centroids with differing texture characteristics are more likely to connect, consistent with a more heterogeneous lesion architecture.
Similarly, Energy Modularity (Figure 5e; $p=0.012$)—a metric of how cleanly a network can be partitioned into distinct modules \cite{Rubinov2010}—was elevated in PsP and reduced in TP. Higher modularity in pseudo-progression indicates stronger subnetwork organization, whereas the lower values in true progression could imply that the infiltrative tumor growth may disrupt clear graph communities, making module delineation more difficult.
Finally, the Homogeneity Clustering Coefficient (Figure 5g; $p=0.008$)—which quantifies the degree of local interconnectedness among nodes sharing similar attributes \cite{Watts1998}—was greater in PsP cases. This may suggest that treatment-affected regions form compact clusters of similar voxel intensities, while the more irregular and invasive growth patterns in TP lesions may lead to reduced local clustering.

Our results demonstrated that GrRAiL achieved a CV accuracy of $89\%$, a test accuracy of $78\%$, and an AUC value of $0.85$ in classifying true progression from pseudo-progression in glioblastoma studies. While patel et al. \cite{patel2021machine} focused on radiomic features from the multiparametric MRI scans and reported an AUC of $0.80$, our approach combined graph features from all 13 radiomic maps, which significantly enhanced performance metrics. Specifically, combining these features increased the AUC to $0.85$ in glioblastoma studies and an AUC of $0.86$ in metastatic brain tumors. This indicates that integrating multiple radiomic-derived graph features may provide a more comprehensive characterization of tumor heterogeneity and improve the distinction of TuR from PsP. Other studies employing graph-theory approaches, such as Cao et al. \cite{cao2023multidimensional}, developed a multidimensional framework combining clinical factors, connectomics, and radiomics features to distinguish tumor recurrence from treatment effects, achieving promising results with an AUC of $0.89$. In their approach, the tumor network was modeled by treating each voxel as a node and applying an empirical threshold of $0.1$ to reduce unnecessary connections in the graph. However, this thresholding may have led to the loss of information by omitting weaker but potentially significant connections within the tumor network.  
Additionally, their model was trained and evaluated using leave-one-out cross-validation without testing on cross-institutional data, which may limit the generalizability of their findings. Lee et al. \cite{lee2023association} explored the association of graph-based spatial features with the overall survival status of glioblastoma patients. They grouped the tumor region into high and low-intensity habitats based on MRI scans and constructed graphs by connecting these regions, focusing primarily on the center slice of the tumor for graph construction. By extracting graph features such as the minimum spanning tree (MST) and gray-level run-length matrix (GRLM), they achieved AUC values of $0.83$ for MST and $0.77$ for GRLM in predicting overall survival. While their study demonstrated the utility of graph-based features in a prognostic application, it was limited to a two-dimensional analysis of a single slice, which may not capture the full complexity of tumor heterogeneity.

\subsection{Distinguishing Tumor Recurrence versus Radiation Necrosis in Metastatic Brain Tumors}

Our metastatic brain tumor graphs showed that the Entropy Average Path Length (Figure 8d, $p < 0.0001$) and the Entropy Diameter (Figure 8e, $p < 0.0001$) were both greater in the TP cohort than in the RN cohort. The average path length—calculated as the mean of all shortest‐path distances between cluster‐centroid nodes—was higher in TuR lesions, which may indicate that, on average, more graph‐steps are required to traverse between typical texture clusters. The diameter, defined as the single longest shortest‐path in the graph, was also larger in TuR, suggesting that the most distant pair of clusters in TuR lesions could be separated by a wider network span than in RN lesions. Together, these two measures—mean versus maximum shortest‐path distance—may reflect a more dispersed arrangement of texture clusters in TP compared to the relatively compact network topology observed in RN \cite{Bartolomei2006}.

Overall, GrRAiL demonstrated a descent performance in distinguishing tumor recurrence (TuR) from radiation necrosis (RN) in metastatic brain tumor cases, achieving a cross-validation accuracy of 84\%, an external test accuracy of 74\%, and an AUC of 0.86. Importantly, GrRAiL was validated using a multi-institutional dataset consisting of 332 lesions from 233 patients, underscoring its clinical relevance and potential generalizability. By contrast, many prior studies on this clinical issue have been single-institutional with limited external validation \cite{bhandari2022machine}. For instance, Basree et al. \cite{basree2024leveraging} reported an AUC of 0.76 using conventional radiomic features from a single-center cohort. Additionally, \cite{gao2020differentiation} employed support vector machine (SVM) classifiers based on radiomic features from pre- and post-contrast T1-weighted and T2-FLAIR subtraction images in a two-center study involving 56 glioma patients, achieving a test accuracy accuracy of 93.3\%, and an AUC of 0.94. While these results appear impressive, \cite{gao2020differentiation} study faced several limitations, including a relatively small sample size, and the absence of external validation cohorts. In contrast, our approach not only leveraged advanced graph-based metrics to capture spatial heterogeneity but also benefited from a substantially larger dataset and external validation across multiple institutions.

\subsection{No-Risk + Low-Risk versus High-Risk Pancreatic Intraductal 
Papillary Mucinous Neoplasm Lesions}
Contrast diameter, sum entropy number of hubs, and energy network resilience emerged as the top‐performing GrRAiL features to discriminate low‐ versus high‐risk IPMNs (p = 0.00022, 0.0024, and 0.008, respectively). The box plot for contrast diameter shows higher values in the high‐risk group and lower values in the low‐risk group, which may reflect greater heterogeneity and spatial dispersion of texture clusters in high‐risk lesions. Similarly, a markedly larger sum entropy number of hubs was observed in high‐risk cases, suggesting a more heterogeneous and interconnected network structure, whereas low‐risk lesions tended to exhibit fewer hubs and potentially simpler topology. In contrast, energy network resilience was higher in the low‐risk class (Figure 11e), which could indicate more uniform or robust connectivity patterns compared with the reduced resilience seen in high‐risk lesions. Notably, contrast‐derived graph features consistently ranked among the strongest discriminators across all three use cases in our study, suggesting they may effectively capture key aspects of tumor heterogeneity and its spatial organization.

Our GrRAiL framework achieved robust performance, demonstrating a cross-validation accuracy of 85\%, external test accuracy of 79\%, and an AUC of 0.83. Notably, GrRAiL was validated using a significantly larger and heterogeneous multi-institutional dataset (n=609), underlining its strong generalizability and applicability to real-world clinical scenarios. In contrast, \cite{chakraborty2018ct} utilized conventional radiomic features extracted from CT scans of a comparatively smaller, single-institution cohort (n=103), obtaining an internal validation mean AUC of 0.77, improving slightly to 0.81 with the addition of clinical variables. Although \cite{cui2021radiomic} reported higher external validation AUCs (0.88 and 0.87) for predicting pathological grades of IPMN lesions using MRI radiomics, their analysis was constrained by a smaller dataset (n=202), manual segmentation of lesions, exclusion of peritumoral regions, and fewer clinical parameters, which potentially limit scalability and broader generalizability. Thus, while \cite{cui2021radiomic} reported slightly higher AUC values, GrRAiL may offer stronger overall performance in terms of dataset size, generalizability, external validation rigor, and the integration of advanced graph-based metrics, potentially enhancing its clinical reliability and practical utility in predicting high-risk lesions.

\subsection{Future Directions}
Although GrRAiL consistently demonstrated improved diagnostic performance across different use-cases in brain and pancreatic neoplasms, there is still
room for improvement. For instance, in the current study, we used only T1-post contrast image sequences to develop GrRAiL for brain tumor studies, and T2w image sequences for IPMN lesions. Extracting features from multimodal sequences, including T1w, T2w, and FLAIR image sequences, might further improve classification accuracies of our trained models. 
Future work will also aim to expand the application of GrRAiL to other cancer types and integrate multi-parametric imaging data to improve its diagnostic performance. Additionally, we aim to explore advanced graph neural network architectures to further improve the performance of the graph-generated adjacency matrices. While we employed multi-institutional studies across different experimental validations of GrRAiL, we intend to expand our study pool to demonstrate consistent (and improved) performance of the GrRAiL descriptor across data from multiple institutions as well as across different disease domains in the future.

\section{CONCLUSION}

In this study, we presented and evaluated the Graph-Radiomic Learning (GrRAiL) descriptor for distinguishing benign confounding pathologies from malignant neoplasms on clinical MRI scans. By leveraging graph representations of radiomic feature maps, GrRAiL seeks to capture the intricate spatial relationships within the lesion microenvironment, to be able to reliably distinguish confounding pathologies from malignant neoplasms.
Our multi-institutional analysis across different use cases encompassing glioblastoma, brain metastasis, and IPMN lesions cohorts demonstrated that  GrRAiL may serve as a surrogate image-based biomarker for distinguishing between recurrent brain tumors and radiation-induced treatment effects, as well as malignancies with no+low versus high-risk, outperforming conventional radiomic and SOTA graph-neural network-based approaches, on clinical MRI scans.

\section{DATA AVAILABILITY}

The MRI datasets for the three study cohorts—glioblastoma, metastatic brain tumors, and intraductal papillary mucinous neoplasms (IPMN)—were acquired under institutional approvals at multiple participating sites. Because of institutional and patient-privacy protections, raw imaging and associated clinical data cannot be publicly released. Upon reasonable request, a data sharing agreement can be initiated between the interested parties and the clinical in- stitution following institution-specific guidelines.

\section{CODE AVAILABILITY}

The underlying code for this study is not publicly available but may be made available to qualified researchers on reasonable request from the corresponding author.

\section{ACKNOWLEDGMENT}
The authors would like to thank Virginia Hill, Volodymyr Statsevych, Raymond Huang, Andrew Baschnagel, Alan McMillan, Michael Veronesi, and Ankush Bhatia for providing access to anonymized brain tumor MRI scans used in this study. 

\section{FUNDING}

This research was supported by NIH/NCI/ITCR (1U01CA248226-01), NIH/NCI (R01CA277728-01A1), NIH/NCI (1R01CA264017-01A1), NIH/NCI (3U01CA248226-03S1), the DOD/PRCRP Career Development Award (W81XWH-18-1-0404), the Dana Foundation David Mahoney Neuroimaging Program, the V Foundation Translational Research Award, the Johnson \& Johnson WiSTEM2D Award, the Musella Foundation Grant, the R \& D Pilot Award, the Departments of Radiology and Medical Physics at the University of Wisconsin--Madison, and the WARF Accelerator Oncology Diagnostics Award.

\section{AUTHOR CONTRIBUTIONS}

Dheerendranath Battalapalli: Data curation, formal analysis, investigation, visualization, methodology, algorithm development, validation, and writing-original draft.
Marwa Ismail: Formal analysis, project administration, review and editing of the manuscript.
Apoorva Safai: Review and editing of the manuscript.
Hyemin Um, Maria Jaramillo, Gustavo Adalfo Pineda Ortiz: Data preprocessing.
Ulas Bagci, Manmeet Singh Ahluwalia: Data provision and review of the manuscript.
Pallavi Tiwari: Conceptualization, funding acquisition, project administration, review and editing of the manuscript.

\section{CORRESPONDING AUTHOR}

Pallavi Tiwari  (e-mail: ptiwari9@wisc.edu)

\section{ETHICS APPROVAL}

This study was approved by the University of Wisconsin–Madison Institutional Review Board (Approval No. 2023-0057 for the Glioblastoma and Metastasis cohorts; Approval No. 2025-1191 for the IPMN cohort). Given the retrospective design and use of de-identified data, the requirement for informed consent was waived in accordance with institutional policies and HIPAA regulations.

\section{COMPETING INTERESTS}

Dr. Pallavi Tiwari is an equity holder in LivAI Inc. The other authors declare no competing interests.

\section{CLINICAL TRIAL REGISTRATION}

Clinical trial registration number: Not applicable. 

\section{REFERENCES}


\begin{thebibliography}{10}
\providecommand{\url}[1]{#1}
\csname url@samestyle\endcsname
\providecommand{\newblock}{\relax}
\providecommand{\bibinfo}[2]{#2}
\providecommand{\BIBentrySTDinterwordspacing}{\spaceskip=0pt\relax}
\providecommand{\BIBentryALTinterwordstretchfactor}{4}
\providecommand{\BIBentryALTinterwordspacing}{\spaceskip=\fontdimen2\font plus
\BIBentryALTinterwordstretchfactor\fontdimen3\font minus \fontdimen4\font\relax}
\providecommand{\BIBforeignlanguage}[2]{{%
\expandafter\ifx\csname l@#1\endcsname\relax
\typeout{** WARNING: IEEEtran.bst: No hyphenation pattern has been}%
\typeout{** loaded for the language `#1'. Using the pattern for}%
\typeout{** the default language instead.}%
\else
\language=\csname l@#1\endcsname
\fi
#2}}
\providecommand{\BIBdecl}{\relax}
\BIBdecl

\bibitem{verhaak2010integrated}
R.~G. Verhaak, K.~A. Hoadley, E.~Purdom, V.~Wang, Y.~Qi, M.~D. Wilkerson, C.~R. Miller, L.~Ding, T.~Golub, J.~P. Mesirov \emph{et~al.}, ``Integrated genomic analysis identifies clinically relevant subtypes of glioblastoma characterized by abnormalities in pdgfra, idh1, egfr, and nf1,'' \emph{Cancer cell}, vol.~17, no.~1, pp. 98--110, 2010.

\bibitem{zhou2022treatment}
Q.~Zhou, C.~Xue, X.~Ke, and J.~Zhou, ``Treatment response and prognosis evaluation in high-grade glioma: An imaging review based on mri,'' \emph{Journal of Magnetic Resonance Imaging}, vol.~56, no.~2, pp. 325--340, 2022.

\bibitem{pope2018brain}
W.~B. Pope, ``Brain metastases: neuroimaging,'' \emph{Handbook of clinical neurology}, vol. 149, pp. 89--112, 2018.

\bibitem{derks2022brain}
S.~H. Derks, A.~A. van~der Veldt, and M.~Smits, ``Brain metastases: the role of clinical imaging,'' \emph{The British journal of radiology}, vol.~95, no. 1130, p. 20210944, 2022.

\bibitem{mitsuya2010perfusion}
K.~Mitsuya, Y.~Nakasu, S.~Horiguchi, H.~Harada, T.~Nishimura, E.~Bando, H.~Okawa, Y.~Furukawa, T.~Hirai, and M.~Endo, ``Perfusion weighted magnetic resonance imaging to distinguish the recurrence of metastatic brain tumors from radiation necrosis after stereotactic radiosurgery,'' \emph{Journal of neuro-oncology}, vol.~99, pp. 81--88, 2010.

\bibitem{mayo2023radiation}
Z.~S. Mayo, A.~Halima, J.~R. Broughman, T.~D. Smile, M.~C. Tom, E.~S. Murphy, J.~H. Suh, S.~S. Lo, G.~H. Barnett, G.~Wu \emph{et~al.}, ``Radiation necrosis or tumor progression? a review of the radiographic modalities used in the diagnosis of cerebral radiation necrosis,'' \emph{Journal of neuro-oncology}, vol. 161, no.~1, pp. 23--31, 2023.

\bibitem{chakraborty2017preliminary}
J.~Chakraborty, L.~Langdon-Embry, K.~M. Cunanan, J.~G. Escalon, P.~J. Allen, M.~A. Lowery, E.~M. O’Reilly, M.~G{\"o}nen, R.~G. Do, and A.~L. Simpson, ``Preliminary study of tumor heterogeneity in imaging predicts two year survival in pancreatic cancer patients,'' \emph{PloS one}, vol.~12, no.~12, p. e0188022, 2017.

\bibitem{gerlinger2012intratumor}
M.~Gerlinger, A.~J. Rowan, S.~Horswell, J.~Larkin, D.~Endesfelder, E.~Gronroos, P.~Martinez, N.~Matthews, A.~Stewart, P.~Tarpey \emph{et~al.}, ``Intratumor heterogeneity and branched evolution revealed by multiregion sequencing,'' \emph{New England journal of medicine}, vol. 366, no.~10, pp. 883--892, 2012.

\bibitem{sottoriva2013intratumor}
A.~Sottoriva, I.~Spiteri, S.~G. Piccirillo, A.~Touloumis, V.~P. Collins, J.~C. Marioni, C.~Curtis, C.~Watts, and S.~Tavar{\'e}, ``Intratumor heterogeneity in human glioblastoma reflects cancer evolutionary dynamics,'' \emph{Proceedings of the National Academy of Sciences}, vol. 110, no.~10, pp. 4009--4014, 2013.

\bibitem{zhou2014radiologically}
M.~Zhou, L.~Hall, D.~Goldgof, R.~Russo, Y.~Balagurunathan, R.~Gillies, and R.~Gatenby, ``Radiologically defined ecological dynamics and clinical outcomes in glioblastoma multiforme: preliminary results,'' \emph{Translational oncology}, vol.~7, no.~1, pp. 5--13, 2014.

\bibitem{zhou2017identifying}
M.~Zhou, B.~Chaudhury, L.~O. Hall, D.~B. Goldgof, R.~J. Gillies, and R.~A. Gatenby, ``Identifying spatial imaging biomarkers of glioblastoma multiforme for survival group prediction,'' \emph{Journal of Magnetic Resonance Imaging}, vol.~46, no.~1, pp. 115--123, 2017.

\bibitem{wu2016robust}
J.~Wu, M.~F. Gensheimer, X.~Dong, D.~L. Rubin, S.~Napel, M.~Diehn, B.~W. Loo~Jr, and R.~Li, ``Robust intratumor partitioning to identify high-risk subregions in lung cancer: a pilot study,'' \emph{International Journal of Radiation Oncology* Biology* Physics}, vol.~95, no.~5, pp. 1504--1512, 2016.

\bibitem{sala2017unravelling}
E.~Sala, E.~Mema, Y.~Himoto, H.~Veeraraghavan, J.~Brenton, A.~Snyder, B.~Weigelt, and H.~Vargas, ``Unravelling tumour heterogeneity using next-generation imaging: radiomics, radiogenomics, and habitat imaging,'' \emph{Clinical radiology}, vol.~72, no.~1, pp. 3--10, 2017.

\bibitem{zhou2018radiomics}
M.~Zhou, J.~Scott, B.~Chaudhury, L.~Hall, D.~Goldgof, K.~W. Yeom, M.~Iv, Y.~Ou, J.~Kalpathy-Cramer, S.~Napel \emph{et~al.}, ``Radiomics in brain tumor: image assessment, quantitative feature descriptors, and machine-learning approaches,'' \emph{American Journal of Neuroradiology}, vol.~39, no.~2, pp. 208--216, 2018.

\bibitem{corral2019deep}
J.~E. Corral, S.~Hussein, P.~Kandel, C.~W. Bolan, U.~Bagci, and M.~B. Wallace, ``Deep learning to classify intraductal papillary mucinous neoplasms using magnetic resonance imaging,'' \emph{Pancreas}, vol.~48, no.~6, pp. 805--810, 2019.

\bibitem{lalonde2019inn}
R.~LaLonde, I.~Tanner, K.~Nikiforaki, G.~Z. Papadakis, P.~Kandel, C.~W. Bolan, M.~B. Wallace, and U.~Bagci, ``Inn: inflated neural networks for ipmn diagnosis,'' in \emph{International Conference on Medical Image Computing and Computer-Assisted Intervention}.\hskip 1em plus 0.5em minus 0.4em\relax Springer, 2019, pp. 101--109.

\bibitem{hussein2019lung}
S.~Hussein, P.~Kandel, C.~W. Bolan, M.~B. Wallace, and U.~Bagci, ``Lung and pancreatic tumor characterization in the deep learning era: novel supervised and unsupervised learning approaches,'' \emph{IEEE transactions on medical imaging}, vol.~38, no.~8, pp. 1777--1787, 2019.

\bibitem{van2020radiomics}
J.~E. Van~Timmeren, D.~Cester, S.~Tanadini-Lang, H.~Alkadhi, and B.~Baessler, ``Radiomics in medical imaging—“how-to” guide and critical reflection,'' \emph{Insights into imaging}, vol.~11, no.~1, p.~91, 2020.

\bibitem{horvat2018mr}
N.~Horvat, H.~Veeraraghavan, M.~Khan, I.~Blazic, J.~Zheng, M.~Capanu, E.~Sala, J.~Garcia-Aguilar, M.~J. Gollub, and I.~Petkovska, ``Mr imaging of rectal cancer: radiomics analysis to assess treatment response after neoadjuvant therapy,'' \emph{Radiology}, vol. 287, no.~3, pp. 833--843, 2018.

\bibitem{mccague2023introduction}
C.~McCague, S.~Ramlee, M.~Reinius, I.~Selby, D.~Hulse, P.~Piyatissa, V.~Bura, M.~Crispin-Ortuzar, E.~Sala, and R.~Woitek, ``Introduction to radiomics for a clinical audience,'' \emph{Clinical Radiology}, vol.~78, no.~2, pp. 83--98, 2023.

\bibitem{ismail2022radiomic}
M.~Ismail, P.~Prasanna, K.~Bera, V.~Statsevych, V.~Hill, G.~Singh, S.~Partovi, N.~Beig, S.~McGarry, P.~Laviolette \emph{et~al.}, ``Radiomic deformation and textural heterogeneity (r-depth) descriptor to characterize tumor field effect: Application to survival prediction in glioblastoma,'' \emph{IEEE transactions on medical imaging}, vol.~41, no.~7, pp. 1764--1777, 2022.

\bibitem{ismail2018shape}
M.~Ismail, V.~Hill, V.~Statsevych, R.~Huang, P.~Prasanna, R.~Correa, G.~Singh, K.~Bera, N.~Beig, R.~Thawani \emph{et~al.}, ``Shape features of the lesion habitat to differentiate brain tumor progression from pseudoprogression on routine multiparametric mri: a multisite study,'' \emph{American Journal of Neuroradiology}, vol.~39, no.~12, pp. 2187--2193, 2018.

\bibitem{battalapalli2023fractal}
D.~Battalapalli, S.~Vidyadharan, B.~Prabhakar~Rao, P.~Yogeeswari, C.~Kesavadas, and V.~Rajagopalan, ``Fractal dimension: analyzing its potential as a neuroimaging biomarker for brain tumor diagnosis using machine learning,'' \emph{Frontiers in Physiology}, vol.~14, p. 1201617, 2023.

\bibitem{yang2015discrete}
G.~Yang, T.~Nawaz, T.~R. Barrick, F.~A. Howe, and G.~Slabaugh, ``Discrete wavelet transform-based whole-spectral and subspectral analysis for improved brain tumor clustering using single voxel mr spectroscopy,'' \emph{IEEE Transactions on Biomedical Engineering}, vol.~62, no.~12, pp. 2860--2866, 2015.

\bibitem{lambin2012radiomics}
P.~Lambin, E.~Rios-Velazquez, R.~Leijenaar, S.~Carvalho, R.~G. Van~Stiphout, P.~Granton, C.~M. Zegers, R.~Gillies, R.~Boellard, A.~Dekker \emph{et~al.}, ``Radiomics: extracting more information from medical images using advanced feature analysis,'' \emph{European journal of cancer}, vol.~48, no.~4, pp. 441--446, 2012.

\bibitem{a2016polypharmacology}
A.~A~Antolin, P.~Workman, J.~Mestres, and B.~Al-Lazikani, ``Polypharmacology in precision oncology: current applications and future prospects,'' \emph{Current pharmaceutical design}, vol.~22, no.~46, pp. 6935--6945, 2016.

\bibitem{aerts2014decoding}
H.~Aerts, E.~Velazquez, R.~Leijenaar, C.~Parmar, P.~Grossmann, S.~Carvalho, J.~Bussink, R.~Monshouwer, B.~Haibe-Kains, D.~Rietveld \emph{et~al.}, ``Decoding tumour phenotype by noninvasive imaging using a quantitative radiomics approach. nat commun. 2014; 5: 4006,'' 2014.

\bibitem{ibrahim2021radiomics}
A.~Ibrahim, S.~Primakov, M.~Beuque, H.~Woodruff, I.~Halilaj, G.~Wu, T.~Refaee, R.~Granzier, Y.~Widaatalla, R.~Hustinx \emph{et~al.}, ``Radiomics for precision medicine: Current challenges, future prospects, and the proposal of a new framework,'' \emph{Methods}, vol. 188, pp. 20--29, 2021.

\bibitem{antunes2022radiomic}
J.~T. Antunes, M.~Ismail, I.~Hossain, Z.~Wang, P.~Prasanna, A.~Madabhushi, P.~Tiwari, and S.~E. Viswanath, ``Radiomic spatial textural descriptor (radistat): Quantifying spatial organization of imaging heterogeneity associated with tumor response to treatment,'' \emph{IEEE journal of biomedical and health informatics}, vol.~26, no.~6, pp. 2627--2636, 2022.

\bibitem{lee2023association}
J.~Lee, S.~Narang, J.~Martinez, G.~Rao, and A.~Rao, ``Association of graph-based spatial features with overall survival status of glioblastoma patients,'' \emph{Scientific Reports}, vol.~13, no.~1, p. 17046, 2023.

\bibitem{cao2023multidimensional}
Y.~Cao, V.~S. Parekh, E.~Lee, X.~Chen, K.~J. Redmond, J.~J. Pillai, L.~Peng, M.~A. Jacobs, and L.~R. Kleinberg, ``A multidimensional connectomics-and radiomics-based advanced machine-learning framework to distinguish radiation necrosis from true progression in brain metastases,'' \emph{Cancers}, vol.~15, no.~16, p. 4113, 2023.

\bibitem{kocevar2016graph}
G.~Kocevar, C.~Stamile, S.~Hannoun, F.~Cotton, S.~Vukusic, F.~Durand-Dubief, and D.~Sappey-Marinier, ``Graph theory-based brain connectivity for automatic classification of multiple sclerosis clinical courses,'' \emph{Frontiers in neuroscience}, vol.~10, p. 478, 2016.

\bibitem{Battalapalli2024}
D.~Battalapalli, A.~Safai, M.~Ismail, V.~Hill, V.~Statsevych, R.~Huang, M.~S. Ahluwalia, and P.~Tiwari, ``Graph-radiomics learning (grrail): Application to distinguishing recurrence from pseudo-progression on structural mri,'' in \emph{Proceedings of the IEEE International Symposium on Biomedical Imaging (ISBI)}, 2024.

\bibitem{haralick1973textural}
R.~M. Haralick, K.~Shanmugam, and I.~H. Dinstein, ``Textural features for image classification,'' \emph{IEEE Transactions on systems, man, and cybernetics}, no.~6, pp. 610--621, 1973.

\bibitem{zhao2021anomaly}
H.~Zhao, Y.~Li, N.~He, K.~Ma, L.~Fang, H.~Li, and Y.~Zheng, ``Anomaly detection for medical images using self-supervised and translation-consistent features,'' \emph{IEEE Transactions on Medical Imaging}, vol.~40, no.~12, pp. 3641--3651, 2021.

\bibitem{zhang2022deepemd}
C.~Zhang, Y.~Cai, G.~Lin, and C.~Shen, ``Deepemd: Differentiable earth mover's distance for few-shot learning,'' \emph{IEEE Transactions on Pattern Analysis and Machine Intelligence}, vol.~45, no.~5, pp. 5632--5648, 2022.

\bibitem{stupp2005radiotherapy}
R.~Stupp, W.~P. Mason, M.~J. Van Den~Bent, M.~Weller, B.~Fisher, M.~J. Taphoorn, K.~Belanger, A.~A. Brandes, C.~Marosi, U.~Bogdahn \emph{et~al.}, ``Radiotherapy plus concomitant and adjuvant temozolomide for glioblastoma,'' \emph{New England journal of medicine}, vol. 352, no.~10, pp. 987--996, 2005.

\bibitem{gondi2022radiation}
V.~Gondi, G.~Bauman, L.~Bradfield, S.~H. Burri, A.~R. Cabrera, D.~A. Cunningham, B.~R. Eaton, J.~A. Hattangadi-Gluth, M.~M. Kim, R.~Kotecha \emph{et~al.}, ``Radiation therapy for brain metastases: an astro clinical practice guideline,'' \emph{Practical radiation oncology}, vol.~12, no.~4, pp. 265--282, 2022.

\bibitem{zhang2025large}
Z.~Zhang, E.~Keles, G.~Durak, Y.~Taktak, O.~Susladkar, V.~Gorade, D.~Jha, A.~C. Ormeci, A.~Medetalibeyoglu, L.~Yao \emph{et~al.}, ``Large-scale multi-center ct and mri segmentation of pancreas with deep learning,'' \emph{Medical image analysis}, vol.~99, p. 103382, 2025.

\bibitem{yushkevich2016itk}
P.~A. Yushkevich, Y.~Gao, and G.~Gerig, ``Itk-snap: An interactive tool for semi-automatic segmentation of multi-modality biomedical images,'' in \emph{2016 38th annual international conference of the IEEE engineering in medicine and biology society (EMBC)}.\hskip 1em plus 0.5em minus 0.4em\relax IEEE, 2016, pp. 3342--3345.

\bibitem{tustison2010n4itk}
N.~J. Tustison, B.~B. Avants, P.~A. Cook, Y.~Zheng, A.~Egan, P.~A. Yushkevich, and J.~C. Gee, ``N4itk: improved n3 bias correction,'' \emph{IEEE transactions on medical imaging}, vol.~29, no.~6, pp. 1310--1320, 2010.

\bibitem{prasanna2017radiomic}
P.~Prasanna, J.~Patel, S.~Partovi, A.~Madabhushi, and P.~Tiwari, ``Radiomic features from the peritumoral brain parenchyma on treatment-naive multi-parametric mr imaging predict long versus short-term survival in glioblastoma multiforme: preliminary findings,'' \emph{European radiology}, vol.~27, pp. 4188--4197, 2017.

\bibitem{guyon2002gene}
I.~Guyon, J.~Weston, S.~Barnhill, and V.~Vapnik, ``Gene selection for cancer classification using support vector machines,'' \emph{Machine learning}, vol.~46, pp. 389--422, 2002.

\bibitem{lundberg2017unified}
S.~Lundberg, ``A unified approach to interpreting model predictions,'' \emph{arXiv preprint arXiv:1705.07874}, 2017.

\bibitem{Hamilton2017GraphSAGE}
W.~L. Hamilton, R.~Ying, and J.~Leskovec, ``Inductive representation learning on large graphs,'' in \emph{Advances in Neural Information Processing Systems (NeurIPS)}, 2017, pp. 1024--1034.

\bibitem{Xu2018GIN}
K.~Xu, W.~Hu, J.~Leskovec, and S.~Jegelka, ``How powerful are graph neural networks?'' in \emph{International Conference on Learning Representations (ICLR)}, 2019.

\bibitem{Velickovic2018GAT}
P.~Veli{\v{c}}kovi{\'{c}}, G.~Cucurull, A.~Casanova, A.~Romero, P.~Li{\`{o}}, and Y.~Bengio, ``Graph attention networks,'' in \emph{International Conference on Learning Representations (ICLR)}, 2018.

\bibitem{Xu2018GCNJK}
K.~Xu, C.~Li, Y.~Bengio, J.~Lee, T.~L. Willke, J.~Pastor-Pellicer, L.~Perron, and S.~Jegelka, ``Representation learning on graphs with jumping knowledge networks,'' in \emph{International Conference on Machine Learning (ICML)}, 2018.

\bibitem{gillies2016radiomics}
R.~J. Gillies, P.~E. Kinahan, and H.~Hricak, ``Radiomics: images are more than pictures, they are data,'' \emph{Radiology}, vol. 278, no.~2, pp. 563--577, 2016.

\bibitem{VanGriethuysen2017}
J.~Van~Griethuysen, A.~Fedorov, C.~Parmar, A.~Hosny, N.~Aucoin, V.~Narayan, R.~Beets-Tan, J.~Fillion-Robin, S.~Pieper, and H.~Aerts, ``Computational radiomics system to decode the radiographic phenotype,'' \emph{Cancer Research}, vol.~77, no.~21, pp. e104--e107, 2017.

\bibitem{Battalapalli2024SIGL}
D.~Battalapalli, A.~Safai, M.~Ismail, V.~Hill, V.~Statsevych, R.~Huang, and P.~Tiwari, ``Spatial interactions via graph-based learning (sigl) to distinguish glioblastoma recurrence from pseudo-progression on clinical mri,'' in \emph{Medical Imaging 2024: Computer-Aided Diagnosis}, ser. Proceedings of SPIE, vol. 12927.\hskip 1em plus 0.5em minus 0.4em\relax SPIE, April 2024, pp. 256--261.

\bibitem{Bullmore2009}
E.~Bullmore and O.~Sporns, ``Complex brain networks: graph theoretical analysis of structural and functional systems,'' \emph{Nature Reviews Neuroscience}, vol.~10, no.~3, pp. 186--198, 2009.

\bibitem{Rubinov2010}
M.~Rubinov and O.~Sporns, ``Complex network measures of brain connectivity: uses and interpretations,'' \emph{Neuroimage}, vol.~52, no.~3, pp. 1059--1069, 2010.

\bibitem{Watts1998}
D.~J. Watts and S.~H. Strogatz, ``Collective dynamics of 'small-world' networks,'' \emph{Nature}, vol. 393, no. 6684, pp. 440--442, 1998.

\bibitem{patel2021machine}
M.~Patel, J.~Zhan, K.~Natarajan, R.~Flintham, N.~Davies, P.~Sanghera, J.~Grist, V.~Duddalwar, A.~Peet, and V.~Sawlani, ``Machine learning-based radiomic evaluation of treatment response prediction in glioblastoma,'' \emph{Clinical radiology}, vol.~76, no.~8, pp. 628--e17, 2021.

\bibitem{Bartolomei2006}
F.~Bartolomei, I.~Bosma, M.~Klein, J.~C. Baayen, J.~C. Reijneveld, T.~J. Postma, and C.~J. Stam, ``Disturbed functional connectivity in brain tumour patients: Evaluation by graph analysis of synchronization matrices,'' \emph{Clinical Neurophysiology}, vol. 117, no.~9, pp. 2039--2049, 2006.

\bibitem{bhandari2022machine}
A.~Bhandari, R.~Marwah, J.~Smith, D.~Nguyen, A.~Bhatti, C.~P. Lim, and A.~Lasocki, ``Machine learning imaging applications in the differentiation of true tumour progression from treatment-related effects in brain tumours: a systematic review and meta-analysis,'' \emph{Journal of Medical Imaging and Radiation Oncology}, vol.~66, no.~6, pp. 781--797, 2022.

\bibitem{basree2024leveraging}
M.~M. Basree, C.~Li, H.~Um, A.~H. Bui, M.~Liu, A.~Ahmed, P.~Tiwari, A.~B. McMillan, and A.~M. Baschnagel, ``Leveraging radiomics and machine learning to differentiate radiation necrosis from recurrence in patients with brain metastases,'' \emph{Journal of Neuro-Oncology}, pp. 1--10, 2024.

\bibitem{gao2020differentiation}
X.-Y. Gao, Y.-D. Wang, S.-M. Wu, W.-T. Rui, D.-N. Ma, Y.~Duan, A.-N. Zhang, Z.-W. Yao, G.~Yang, and Y.-P. Yu, ``Differentiation of treatment-related effects from glioma recurrence using machine learning classifiers based upon pre-and post-contrast t1wi and t2 flair subtraction features: a two-center study,'' \emph{Cancer management and research}, pp. 3191--3201, 2020.

\bibitem{chakraborty2018ct}
J.~Chakraborty, A.~Midya, L.~Gazit, M.~Attiyeh, L.~Langdon-Embry, P.~J. Allen, R.~K. Do, and A.~L. Simpson, ``Ct radiomics to predict high-risk intraductal papillary mucinous neoplasms of the pancreas,'' \emph{Medical physics}, vol.~45, no.~11, pp. 5019--5029, 2018.

\bibitem{cui2021radiomic}
S.~Cui, T.~Tang, Q.~Su, Y.~Wang, Z.~Shu, W.~Yang, and X.~Gong, ``Radiomic nomogram based on mri to predict grade of branching type intraductal papillary mucinous neoplasms of the pancreas: A multicenter study,'' \emph{Cancer Imaging}, vol.~21, pp. 1--13, 2021.

\end{thebibliography}


\end{document}